\documentclass{article}

\usepackage{arxiv}

\usepackage[utf8]{inputenc} 
\usepackage[T1]{fontenc}    
\usepackage{hyperref}       
\usepackage{url}            
\usepackage{booktabs}       
\usepackage{amsfonts}       
\usepackage{nicefrac}       
\usepackage{microtype}      
\usepackage{lipsum}		
\usepackage{graphicx}
\usepackage{doi}
\usepackage{xcolor}
\usepackage{svg}
\usepackage{subcaption} 
\usepackage{booktabs}
\usepackage{siunitx}
\usepackage{geometry}
\usepackage{graphicx}

\title{Shearlet Neural Operators for Anisotropic-Shock-Dominated and Multi-scale parametric partial differential equations}

\author{ 
\href{https://orcid.org/0000-0002-9650-3619
}{\includegraphics[scale=0.06]{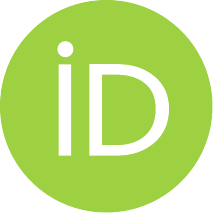}
\hspace{1mm}Fábio Pereira dos Santos}\\
Coordination of Mathematical and Computational Methods\\
National Laboratory of Scientific Computing\\
Av. Getúlio Vargas, 333 - Quitandinha, Petrópolis - RJ, 25651-075 \\
\texttt{fpsantos@lncc.br} \\
\And
\href{https://orcid.org/0000-0001-5472-3544
}{\includegraphics[scale=0.06]{orcid.pdf}
\hspace{1mm}Júlio de Castro Vargas Fernandes}\\
Coordination of Mathematical and Computational Methods\\
National Laboratory of Scientific Computing\\
Av. Getúlio Vargas, 333 - Quitandinha, Petrópolis - RJ, 25651-075 \\
\texttt{julio@lncc.br} \\
\And
\href{https://orcid.org/0000-0002-2941-2522
}{\includegraphics[scale=0.06]{orcid.pdf}
\hspace{1mm}Adriano Maurício de Almeida Côrtes}\\
Systems and Computational Engineering Program, COPPE/UFRJ,\\
Federal University of Rio de Janeiro,\\ P.O. Box 68542, Rio de Janeiro, RJ, 21941-909, Brasil
\texttt{adricortes@cos.ufrj.br} \\
}

\renewcommand{\shorttitle}{\textit{arXiv} Template}

\begin{document}
\maketitle

\begin{abstract}
Neural operators have emerged as powerful data-driven surrogates for learning solution operators of parametric partial differential equations (PDEs). However, widely used Fourier Neural Operators (FNOs) rely on global Fourier representations, which can be inefficient for resolving anisotropic structures, sharp gradients, and spatially localized discontinuities that arise in shock-dominated and multiscale regimes. To address these limitations, we introduce the Shearlet Neural Operator (SNO), a neural operator architecture that replaces the Fourier transform with a shearlet-based representation. Shearlets offer directional, multiscale, and spatially localized atoms with near-optimal sparse approximation of anisotropic features, providing an inductive bias aligned with PDE solutions containing edges, fronts, and shocks. SNO learns in the shearlet domain and reconstructs predictions via the inverse transform, retaining efficient spectral computation while improving locality and directional selectivity. Across seven benchmark PDE families—including strongly anisotropic advection, anisotropic diffusion, and nonlinear conservation laws with straight, curved, interacting, spiral, and polygonal shock structures—SNO consistently improves predictive accuracy and feature fidelity over FNO baselines, with the largest gains observed in anisotropic and discontinuity-dominated settings.
\end{abstract}


\section{Introduction}\label{sec:introduction}
Partial differential equations (PDEs) underpin mathematical models of transport \cite{LeVeque_2002}, diffusion \cite{crank1975mathematics} and wave propagation \cite{LeVeque_2002} \cite{doi:10.1137/1018130} across physics, engineering and the applied sciences. In many contemporary applications--ranging from uncertainty quantification \cite{smith2024uncertainty} \cite{81d297} and inverse problems \cite{isakov2013inverse} to design optimization and real-time decision making--the dominant computational cost is not a single PDE solve, but repeated evaluations across a family of PDEs as coefficients, forcing terms, boundary conditions or geometries vary. Developing fast and reliable surrogates for such parametric PDE families is therefore a central challenge in computational science.

Recent advances in scientific machine learning have led to neural-network-based approaches that approximate PDE solutions, or solution maps, directly from data. Physics-informed neural networks (PINNs) incorporate governing equations into the loss function via automatic differentiation, enabling data-efficient learning for forward and inverse problems \cite{RAISSI2019686}. However, PINNs are typically trained for fixed problem configurations and must be retrained when parameters, including here initial or boundary conditions, change. This limitation has motivated the emergence of operator learning, which aims to approximate resolution-invariant mappings between infinite-dimensional function spaces, namely, from some parametric field to the solution field.

The first representative of operator learning was DeepONet. It introduced a branch–trunk architecture borrowing  ideas from a universal approximation theorem for nonlinear operators \cite{Lu_2021}, as such with nice approximation guarantees. Another well succeeded representative, the Fourier Neural Operator (FNO), demonstrated that departing from the integral representation of the solution, and parameterizing the integral kernels in Fourier space yields efficient, discretization-independent surrogates for parametric PDEs \cite{li2021fourierneuraloperatorparametric}. A unifying theoretical framework later formalized neural operators as mappings between Banach spaces and clarified their approximation, consistency and convergence properties under discretization \cite{10.5555/3648699.3648788}. This perspective has catalyzed a rapidly growing ecosystem of operator-learning architectures, including graph-based operators, low-rank constructions and multigrid-inspired hybrids \cite{HU2025107649}.

Alongside purely data-driven operator learning, hybrid formulations seek to incorporate physical structure to improve robustness and data efficiency. Physics-Informed Neural Operators (PINO) combine supervised learning with PDE-residual regularization, optionally imposed at resolutions finer than the training data \cite{li2023physicsinformedneuraloperatorlearning}. Despite these advances, anisotropic and shock-dominated regimes remain challenging. Many PDEs of scientific and engineering relevance-such as advection-dominated transport, hyperbolic conservation laws and heterogeneous elliptic problems-exhibit sharp fronts, localized discontinuities and directionally organized multiscale features that are difficult to capture with global, isotropic representations.

A key contributing factor is the spectral bias of neural networks: during training, low-frequency components are learned preferentially, while high-frequency structure converges more slowly and less reliably. This phenomenon has been established theoretically \cite{John_Xu_2020,pmlr-v190-luo22a,luo2019theoryfrequencyprinciplegeneral,doi:10.1137/21M1444400,cao2020understandingspectralbiasdeep} and observed empirically across architectures and tasks \cite{rahaman2019spectralbiasneuralnetworks,xu2019trainingbehaviordeepneural,xu2024overviewfrequencyprinciplespectralbias}. In operator-learning settings, spectral bias manifests as oversmoothing and degraded resolution of thin layers, shocks and filamentary structures. Multiscale extensions of Fourier Neural Operators \cite{you2024mscalefnomultiscalefourierneural} and DeepONets \cite{liu2021multiscaledeeponetnonlinearoperators} partially alleviate this issue by introducing hierarchical representations across scales, while wavelet-based operators improve spatial localization. However, commonly used wavelet constructions remain largely isotropic and do not explicitly encode the directional geometry of anisotropic singularities.

In this work, we address these limitations by introducing an architectural prior grounded in shearlet theory. Shearlets form a directional multiscale representation that replaces rotations with shear transformations, enabling faithful digital implementations and near-optimal sparse approximation of functions with anisotropic features such as edges and fronts \cite{kutyniok2006shearlets,kutyniok2012shearlets}. Efficient discrete realizations are well established \cite{Kutyniok_2016}, and shearlets have been explored in deep learning as fixed feature extractors and scattering-type representations \cite{wiatowski2018energypropagationdeepconvolutional,sadiq2020shearletcnn}. To the best of our knowledge, however, shearlets have not been systematically integrated as differentiable spectral layers within neural operator architectures.

We therefore propose the Shearlet Neural Operator (SNO), which embeds shearlet-domain mixing directly into the neural-operator pipeline. By combining global receptive fields and resolution invariance with directionally selective, spatially localized multiscale representations, SNO introduces an inductive bias aligned with the geometry of anisotropic PDE solutions. This design directly targets the representation bottleneck encountered by Fourier-based operators in the presence of shocks, edges and strongly directional fine-scale structure.

Our main contributions are:

\begin{itemize}
    \item A shearlet-based neural operator architecture that introduces directional multiscale inductive bias while preserving efficient spectral computation;
    \item A fully differentiable integration of the digital shearlet transform into operator learning, enabling end-to-end training;
\end{itemize}

The remainder of the paper is organized as follows. Section~\ref{sec:background} reviews neural-operator preliminaries. Section~\ref{sec:methodology} presents the SNO architecture and implementation details. Section~\ref{sec:numerical} describes the benchmark suite, Section~\ref{sec:network_training} details the training protocol, Section~\ref{sec:results} reports quantitative and qualitative results and Section ~\ref{sec:conclusion} details our conclusions.
\section{Background}\label{sec:background}
\paragraph{Notation.}

We denote by $u:\Omega\subset\mathbb{R}^2\to\mathbb{R}^{C_{\mathrm{in}}}$ a function representing a physical field of interest, viewed as an element of a function space $\mathcal{U}$, and by $v:\Omega\to\mathbb{R}^{C_{\mathrm{out}}}$ the corresponding output field. A neural operator is written as $\mathcal{G}_\theta:\mathcal{U}\to\mathcal{V}$, where $\theta$ denotes trainable parameters. In practice, functions $u$ and $v$ are represented on discrete grids, and these discretized representations constitute the inputs and outputs of the neural network. Intermediate feature fields at network layer $\ell$ are denoted by $u_\ell$. The discrete Fourier transform and its inverse are denoted by $\mathcal{F}$ and $\mathcal{F}^{-1}$, respectively, with $\widehat{u}=\mathcal{F}(u)$ evaluated on a discrete frequency grid $(k_x,k_y)$. Scale and shear indices in the shearlet decomposition are denoted by $j$ and $s$; when convenient, a composite index $m\leftrightarrow(j,s)$ is used to simplify notation.

\subsection{Neural Operator Learning}

Neural operators are machine-learning architectures designed to approximate nonlinear operators, that is, mappings between infinite-dimensional function spaces \cite{10.5555/3648699.3648788}. Given Banach (or Hilbert) spaces $\mathcal{U}$ and $\mathcal{V}$, the objective is to approximate a target operator

\[
\mathcal{G} : \mathcal{U} \rightarrow \mathcal{V},
\]

by learning a parameterized neural operator $\mathcal{G}_\theta : \mathcal{U} \rightarrow \mathcal{V}$, with parameters $\theta$, such that $\mathcal{G}_\theta(u) \approx \mathcal{G}(u)$ for all relevant $u \in \mathcal{U}$ in the problem at hand. Here, both the input $u \in \mathcal{U}$ and the output $\mathcal{G}(u) \in \mathcal{V}$ are functions rather than finite-dimensional vectors. Canonical examples include solution operators of parametric partial differential equations, flow maps of dynamical systems, and nonlinear integral transforms.

From a numerical-analysis perspective, many PDE solution operators can be formally expressed in terms of integral operators, such as Green’s functions or resolvents of differential operators. Neural operators can be viewed as data-driven generalizations of these constructions, where the kernel structure and nonlinear corrections are learned directly from data rather than derived analytically. This connection motivates architectures based on nonlocal interactions, which naturally encode long-range dependencies and global constraints ubiquitous in PDEs.

A key distinction from standard neural networks--which typically approximate functions between finite-dimensional Euclidean spaces (e.g., $\mathbb{R}^n \rightarrow \mathbb{R}^m$)--is resolution invariance. Neural operators aim to approximate a continuous mapping that can be evaluated on discretizations of varying resolution and, in some cases, on different meshes. In contrast, conventional convolutional architectures are typically tied to a fixed grid through local receptive fields. From a computational standpoint, operator learning amortizes the cost of repeated PDE solves: once trained, a neural operator enables rapid surrogate evaluations for many parameter instances, making it particularly attractive for inverse problems, uncertainty quantification, and optimization.

Most neural operator architectures approximate the target operator $\mathcal{G}$ through the parameterization of a nonlocal kernel operator, which serves as a building block for the learned operator $\mathcal{G}_\theta$. A generic kernel operator takes the form

\[
\mathcal{K}(u)(y)
=
\int_{\Omega} \kappa(x,y)\,u(x)\,dx,
\]

where $x$ denotes the integration variable and $y$ the evaluation point. The kernel $\kappa:\Omega\times\Omega\to\mathbb{R}$ encodes how information at location $x$ influences the output at location $y$.

Different structural assumptions on $\kappa$ give rise to distinct neural operator formulations. In particular, if the kernel is assumed to be translation invariant, $\kappa(y,x)=\kappa(y-x)$, the operator reduces to a convolution. This structure underlies Fourier Neural Operators, where convolution kernels are parameterized efficiently in the Fourier domain via learned spectral multipliers. More general kernel parameterizations, including wavelet- and graph-based constructions, relax translation invariance to introduce spatial localization or adaptivity to irregular domains. In all cases, the kernel operator is composed with pointwise nonlinearities to capture the nonlinearity of the target operator $\mathcal{G}$. The choice of kernel structure therefore induces a strong inductive bias on the learned operator, motivating the exploration of directional and multiscale parameterizations tailored to anisotropic PDE solutions.

\subsubsection{Fourier Neural Operator (FNO)}

The Fourier Neural Operator (FNO) \cite{li2021fourierneuraloperatorparametric} is a spectral neural operator that approximates the target operator $\mathcal{G}$ by parameterizing a translation-invariant kernel in Fourier space. For an input function $u$, the FNO computes its discrete Fourier transform $\widehat{u}(k_x,k_y)$ and applies a learned complex-valued multiplier on a truncated set of low-frequency modes,

\[
\widehat{v}(k_x,k_y)
=
R(k_x,k_y)\,\widehat{u}(k_x,k_y),
\qquad (k_x,k_y)\in\Omega_k,
\]

where $R(k_x,k_y)$ denotes a trainable tensor of Fourier multipliers and is nonzero only on a fixed rectangular subset of low-frequency modes. This structure corresponds to assuming a translation-invariant convolution kernel. The filtered spectrum is mapped back to physical space through the inverse Fourier transform, yielding a nonlocal update with a global receptive field.

\[
v(x)
=
\mathcal{F}^{-1}\!\left[
\widehat{v}(k_x,k_y)
\right](x),
\]

In practice, the spectral convolution is combined with a pointwise linear map and a nonlinear activation. Writing $u_\ell$ for the feature field at layer $\ell$, a single FNO layer is given by

\[
u_{\ell+1}(x)
=
\sigma\!\left(
\mathcal{F}^{-1}\!\left[
R_\ell(k_x,k_y)\,\widehat{u}_\ell(k_x,k_y)
\right](x)
+
W_\ell u_\ell(x)
\right),
\]

where $W_\ell$ is a learned pointwise ($1\times1$) linear operator and $\sigma$ is a nonlinear activation function. Stacking multiple such layers yields a deep neural operator capable of approximating nonlinear PDE solution operators.

The FNO has become a foundational architecture in operator learning due to (i) its global receptive field induced by spectral convolution, (ii) computational efficiency via FFT-based implementations, and (iii) the ability to generalize across discretizations when inputs and outputs are represented at different resolutions. These properties have enabled strong performance across a range of PDE benchmarks in fluid mechanics, porous-media flow, and related applications.

However, the reliance on global Fourier bases introduces structural limitations. Fourier modes are spatially delocalized and isotropic, making them inefficient for representing sharp fronts, localized discontinuities, and directionally organized features. Truncation to low frequencies further exacerbates this effect, leading to oversmoothing in regimes dominated by anisotropic transport or shock-like behavior. These representational constraints motivate the exploration of alternative spectral parameterizations that retain global coupling while introducing spatial localization and directional sensitivity.

\subsubsection{Limitations of Fourier-Based Representations in Operator Learning}

Despite their computational efficiency and strong performance on smooth problems, Fourier-based representations exhibit intrinsic limitations when applied to PDE solution operators dominated by anisotropy, localization, and discontinuities. These limitations arise from both representational and algorithmic considerations.

First, Fourier modes are globally supported in physical space. As a result, localized features such as shocks, thin layers, or sharp fronts are represented non-sparsely, requiring the superposition of many modes to achieve accurate reconstruction. In practical neural operator implementations, only a finite number of low-frequency modes are retained, inherently biasing the representation toward smooth, globally varying structures and leading to oversmoothing of localized or high-gradient features.

Second, Fourier bases are inherently isotropic and do not encode directional preference. While isotropy is advantageous for problems with homogeneous structure, it becomes a limitation in regimes characterized by strong directional organization, such as advection-dominated transport, anisotropic diffusion, and interacting shock fronts. Directionally aligned features must be represented indirectly through combinations of isotropic modes, reducing representational efficiency.

Third, Fourier representations lack spatial localization: changes in the solution confined to a small region of the domain affect Fourier coefficients globally. This global coupling complicates the modeling of spatially heterogeneous dynamics and limits the ability to adapt representational capacity locally, particularly in parametric PDE families where the location or orientation of sharp features varies across realizations.

A classical manifestation of these limitations is the Gibbs phenomenon, whereby truncated Fourier series exhibit persistent oscillations and overshoots near discontinuities that do not vanish with increasing resolution. In the context of PDE solutions containing shocks or sharp fronts, this behavior reflects a fundamental incompatibility between globally smooth Fourier bases and localized discontinuities. Although neural operators do not explicitly compute Fourier series approximations in the classical sense, the use of truncated Fourier representations induces analogous artifacts, particularly when combined with nonlinearities and spectral filtering.

One natural response to these shortcomings is the use of wavelet representations, which were explicitly developed to address spatial localization and discontinuities through compactly supported, multiscale basis functions. Wavelets provide significantly improved localization compared to Fourier bases and are well suited for representing point singularities and jump discontinuities. However, standard wavelet constructions rely on isotropic scaling and offer only limited directional selectivity. As a consequence, anisotropic features such as edges, ridges, and fronts aligned along arbitrary orientations are not represented sparsely and require many wavelet coefficients to capture their geometry accurately.

This limitation is particularly relevant for PDE solutions in which sharp features are not only localized but also directionally organized, as in anisotropic transport, curved fronts, and interacting shocks. In such settings, spatial localization alone is insufficient: efficient representation additionally requires explicit encoding of orientation and anisotropic scaling. These considerations motivate the exploration of multiscale representations that combine localization with directional sensitivity, while remaining compatible with efficient digital implementations and operator-learning frameworks.

Motivated by the limitations of isotropic Fourier representations, we introduce the Shearlet Neural Operator (SNO), a neural operator architecture that embeds directional multiscale spectral representations inspired by shearlet theory. By replacing the Fourier multiplier in FNOs with a structured collection of anisotropic spectral bands, the SNO preserves global coupling and resolution invariance while introducing an inductive bias aligned with the geometry of anisotropic PDE solutions.
\section{Methodology}\label{sec:methodology}
\subsection{Shearlet Neural Operator}

This section describes the proposed \emph{Shearlet Neural Operator} (SNO), including the construction of discrete shearlet frequency windows, the associated spectral convolution, and the resulting operator architecture. The SNO retains the global coupling and discretization independence of Fourier-based neural operators while introducing a directional multiscale spectral representation tailored to anisotropic PDE solutions.

\subsubsection{Shearlet frequency windows}

Let $u:\Omega\subset\mathbb{R}^2\to\mathbb{R}^{C_{\mathrm{in}}}$ be a function and let

\[
\widehat{u}(k_x,k_y)
=
\mathcal{F}(u)(k_x,k_y)
\]

denote its discrete real Fourier transform evaluated on the corresponding
rFFT frequency grid $\Omega_k\subset\mathbb{R}^2$. Classical shearlet systems are generated via anisotropic parabolic dilations and shear transformations, followed by localization through smooth window functions \cite{kutyniok2006shearlets,kutyniok2012shearlets}. In this work, we construct a digital shearlet-inspired system directly on the discrete Fourier grid, avoiding explicit spatial-domain filtering.

Radial localization is achieved through smooth Meyer-type windows applied to the normalized radial frequency

\[
r(k_x,k_y) = \frac{\sqrt{k_x^2 + k_y^2}}{R_{\max}},
\]

where $R_{\max}$ denotes the maximum radial frequency on the grid. For each scale $j\ge 1$, we define a radial window

\[
\bar{W}_{j}(k_x,k_y)
=
\phi\!\left(
a_j\, r(k_x,k_y)
\right),
\qquad
a_j = 2^{-(j+1)},
\]

where $\phi$ is a smooth Meyer-type taper and $a_j = 2^{-(j+1)}$ controls the dyadic scaling. An additional isotropic low-pass window corresponds to scale index $j=0$. Band-pass behavior at higher scales is obtained by differences of such low-pass windows, while an isotropic low-pass window captures the coarsest frequencies. Directional selectivity is introduced through angular windows defined in polar or slope coordinates. Let $\theta(k_x,k_y) = \arctan2(k_y,k_x)$ denote the angular coordinate in
frequency space. For each scale $j\ge 1$ and shear index $s$, define

\[
A_{j,s}(k_x,k_y)
=
\psi\!\left(
\theta(k_x,k_y) - \theta_{j,s}
\right),
\]

where $\psi$ is a smooth raised-cosine window and $\theta_{j,s} = \arctan(s\,2^{-\lfloor j/2 \rfloor})$ encodes the shear orientation. This parameterization yields a cone-adapted directional tiling of the frequency plane, with the number of orientations increasing with scale to provide finer angular resolution at higher frequencies. 

The resulting shearlet window is defined as the product of radial and angular components,

\begin{equation}
\Xi_{j,s}(k_x,k_y)
=
\bar{W}_j(k_x,k_y)\,A_{j,s}(k_x,k_y).
\end{equation}

An additional isotropic low-pass window corresponds to scale index $j=0$. To ensure numerical stability and balanced energy distribution across scales and orientations, the full collection of windows is normalized to satisfy an approximate tight-frame condition on the discrete frequency grid,

\begin{equation}
\sum_{j,s} \left|\Xi_{j,s}(k_x,k_y)\right|^2 \approx 1,
\qquad \forall (k_x,k_y)\in\Omega_k.
\end{equation}

This normalization is enforced pointwise during construction and ensure that the decomposition acts as a stable partition of unity in frequency space.

\begin{figure}[h!]
    \centering
    \includegraphics[width=\textwidth]{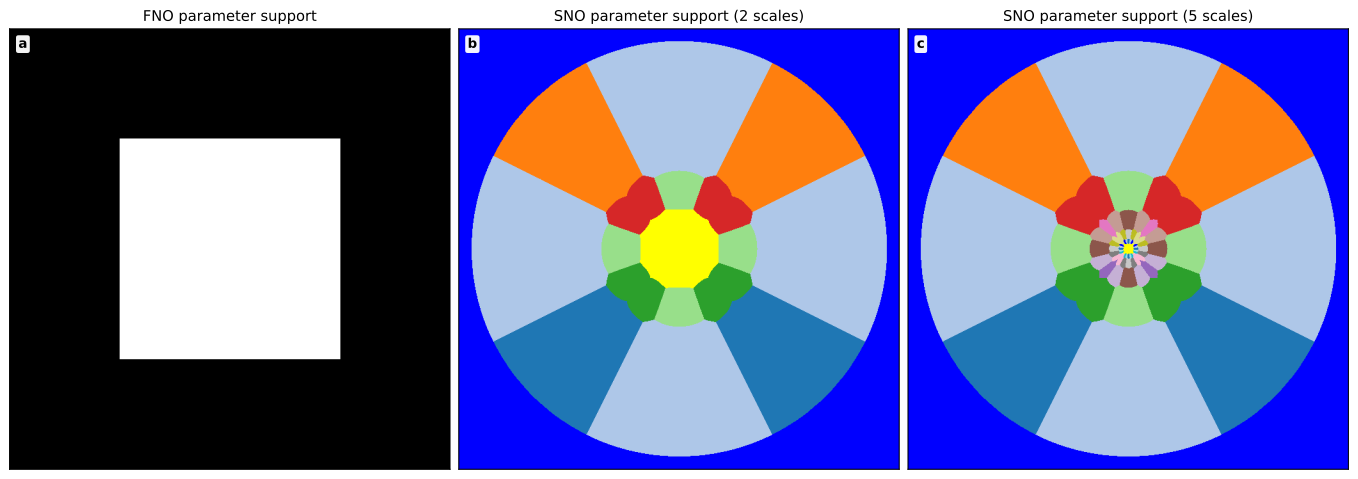}
    \caption{Comparison of spectral parameter support in Fourier and Shearlet Neural Operators. \textbf{(a)} Fourier Neural Operator (FNO): trainable parameters act on a rectangular subset of low-frequency Fourier modes, yielding an isotropic spectral truncation with no directional selectivity. \textbf{(b)} Shearlet Neural Operator (SNO) with two scales: parameters are distributed over wedge-shaped, directionally selective frequency bands, capturing both global structure and coarse directional features. \textbf{(c)} SNO with four scales: increasing the number of scales refines both radial and angular resolution, producing a dense multiscale tiling near the origin and improved directional coverage at higher frequencies. Unlike FNOs, SNOs allocate spectral capacity anisotropically, aligning the learned operator with the geometry of anisotropic and transport-dominated PDE solutions.}
    \label{fig:parameter_support}
\end{figure}

The key distinction between the FNO and SNO lies in this structure of the learnable space where parameters can act. To illustrate this, Figure~\ref{fig:parameter_support} provides a schematic comparison of the spectral supports induced by the two architectures. In the FNO, trainable parameters are associated with a rectangular subset of low-frequency Fourier modes. This isotropic truncation yields a global spectral filter that treats all directions uniformly and allocates capacity primarily to smooth, large-scale components of the solution. High-frequency content outside this rectangular region is implicitly discarded, leading to oversmoothing in regimes dominated by sharp fronts or localized anisotropic features.

In contrast, the SNO distributes trainable parameters across a collection of multiscale, directionally selective shearlet bands. Each band occupies a wedge-shaped region of the frequency plane, corresponding to a specific scale and orientation. Coarse scales capture low-frequency global structure, while finer scales progressively refine angular resolution and extend toward higher frequencies. As illustrated in Figure~\ref{fig:parameter_support}b–c, increasing the number of scales densifies the tiling near the origin while simultaneously improving directional coverage at intermediate and high frequencies. This structured parameter support reflects the underlying design of the SNO layer: instead of learning a single isotropic spectral filter, the operator learns independent complex-valued mixing coefficients within each shearlet band, modulated by scale-dependent gates. As a result, the SNO allocates representational capacity preferentially along directions and scales that align with anisotropic PDE features, while preserving the global coupling characteristic of spectral operator learning.

The figure illustrates parameter support rather than the magnitude of learned coefficients; all spectral filters are learned during training.

\subsubsection{Shearlet spectral convolution}

Let $\{\Xi_{j,s}\}$ denote the collection of shearlet windows indexed by scale $j$ and shear $s$. For notational convenience, we introduce a composite index $m \leftrightarrow (j,s)$ and write $\{\Xi_m\}_{m=0}^{M-1}$ for the full collection.

Let $u_\ell:\Omega\to\mathbb{R}^{C_{\mathrm{in}}}$ denote the feature field at layer $\ell$, and let $u_\ell$ also denote its discretized representation on a grid of size $n_x\times n_y$. Its discrete Fourier transform is denoted by $\widehat{u}_\ell(k_x,k_y)$.

The core operation of the SNO layer is a windowed spectral decomposition,

\begin{equation}
\widehat{u}_{\ell,m}^{(c)}(k_x,k_y)
=
\widehat{u}_\ell^{(c)}(k_x,k_y)\,
\Xi_m(k_x,k_y),
\end{equation}

for each input channel $c=1,\dots,C_{\mathrm{in}}$ and window index $m$. This yields a set of scale- and orientation-resolved spectral components. These components are then linearly combined using learned complex-valued weights,

\begin{equation}
\widehat{v}_\ell^{(o)}(k_x,k_y)
=
\sum_{m=0}^{M-1}
\sum_{c=1}^{C_{\mathrm{in}}}
W_{m,c,o}\,
\widehat{u}_{\ell,m}^{(c)}(k_x,k_y),
\qquad
o=1,\dots,C_{\mathrm{out}},
\end{equation}

where $W_{m,c,o}\in\mathbb{C}$ are trainable spectral mixing coefficients
associated with each shearlet band. To modulate the relative importance of different resolution levels, we introduce a learnable \emph{scale gating} mechanism. Each scale $j$ is assigned a scalar gate $g_j = \sigma(\gamma_j)$ where $\gamma_j$ are trainable parameters and $\sigma$ denotes the sigmoid function. All shearlet windows associated with the same scale share the same gate value. The gated spectral output is given by

\begin{equation}
\widehat{v}_\ell^{(o)}(k_x,k_y)
=
\sum_{m=0}^{M-1}
g_{\mathrm{scale}(m)}
\sum_{c=1}^{C_{\mathrm{in}}}
W_{m,c,o}\,
\widehat{u}_{\ell,m}^{(c)}(k_x,k_y).
\end{equation}

Finally, the spatial output is obtained via the inverse real Fourier transform,

\begin{equation}
v_\ell(x,y)
=
\mathcal{F}^{-1}\!\left[
\widehat{v}_\ell(k_x,k_y)
\right](x,y).
\end{equation}

\subsubsection{SNO block and network architecture}

The SNO layer replaces the Fourier spectral convolution used in Fourier Neural Operators with the shearlet-based spectral convolution described above. As in FNO-type architectures, this spectral operation is combined with a pointwise linear channel mixing and a nonlinear activation. Denoting by $u_\ell:\Omega\to\mathbb{R}^{C}$ the feature field at layer $\ell$, an SNO block is defined as

\begin{equation}
u_{\ell+1}(x,y)
=
\sigma\!\left(
\mathcal{F}^{-1}\!\left[
\widehat{v}_\ell(k_x,k_y)
\right](x,y)
+
W_\ell u_\ell(x,y)
\right),
\end{equation}

where $\widehat{v}_\ell(k_x,k_y)$ denotes the gated shearlet-domain spectral representation obtained in the previous step and $W_\ell$ is a learned $1\times1$ (pointwise) convolution acting on the channel dimension. Stacking multiple such blocks yields the full Shearlet Neural Operator. This architecture preserves the global receptive field and FFT-based efficiency of Fourier Neural Operators while incorporating the geometric adaptivity of shearlets, enabling more efficient representation of anisotropic and directionally organized PDE features.

\subsubsection{Properties and practical advantages}

By combining shearlet-based spectral decomposition with neural operator learning, the SNO inherits several properties that are particularly relevant for PDE surrogate modeling:

\begin{itemize}
    \item \textbf{Directional sensitivity:} shearlet windows decompose the frequency domain into orientation-dependent sectors, enabling more efficient representation of shocks, edges, and ridges.
    \item \textbf{Multiscale representation:} coarse scales capture global structure, while fine scales focus on high-frequency directional features.
    \item \textbf{Global receptive field:} as in FNOs, all spectral modes are processed simultaneously, enabling nonlocal interactions.
    \item \textbf{Learnable anisotropy:} scale gates allow the network to adaptively emphasize or suppress different resolution levels based on the underlying PDE regime.
\end{itemize}

\noindent

In summary, the Shearlet Neural Operator replaces the isotropic Fourier representation used in FNOs with a multiscale directional shearlet system, providing a spectrally structured and geometrically informed neural operator better suited to anisotropic, transport-dominated, and shock-containing PDE dynamics.

\section{Numerical Benchmarks}\label{sec:numerical}
To assess the approximation and generalization performance of the operator-learning models, we consider seven synthetic partial differential equation (PDE) benchmarks. These benchmarks are selected to span a wide range of physical behaviors, encompassing linear and nonlinear advection, anisotropic diffusion, multi-frequency wave interactions, shock formation, spiral wave propagation, and Kelvin–Helmholtz–type shear instabilities. 

\subsection{General Formulation}

Each problem consists of a scalar field \(u(x,y,t)\) defined on a rectangular domain
\[
(x,y)\in [a_x,b_x]\times[a_y,b_y], \qquad t\in [0,T].
\]
The governing PDEs are discretized in space using finite-difference schemes and advanced in time with a third-order Strong Stability Preserving Runge–Kutta (SSP-RK3) method~\cite{GottliebShu1998}, which suppresses spurious oscillations in the presence of shocks. For cases with available analytical solutions, these are employed to prevent the introduction of numerical errors. It is important to note that, for all cases, the stability criterion was strictly observed to avoid any additional sources of error.

\subsection{Benchmark Test Cases}

In this section, We present seven PDE toy-model problems representing a range of physical phenomena, each exhibiting specific directional complexity. For clarity, we classify them into three categories: diffusion-dominated, convection-dominated, and combined-effect problems.

\subsection{Diffusion-dominated}

\subsection{Multi-Orientation Texture}

This toy model represents the superposition of multiple wave components—each with distinct orientations, wavenumbers, and initial phases—propagating through a medium that exhibits anisotropic diffusion. This means that diffusion (or spreading) is direction‑dependent. This type of model is present in some biological systems ~\cite{BEAULIEU2009105}. 

\begin{equation}
\frac{\partial u}{\partial t} = D_x \frac{\partial^2 u}{\partial x^2} + D_y \frac{\partial^2 u}{\partial y^2}, \quad D_x = 0.4, \quad D_y = 12
\end{equation}
\begin{equation}
u_0(x,y) = \sum_{k=1}^{4} \left[\sin\left(10k x + 6k y\right) + \sin\left(12k x - 8k y\right)\right]
\end{equation}

\subsection{Convection-dominated}

\subsubsection{Bent Ridge Advection}

Features a parabolically curved ridge structure, where the ridge orientation varies with $y$. Challenges models to handle spatially varying anisotropy and curved directional features. This meant to describe a front curved advection, such as oceanic fronts~\cite{10.1098/rspa.2016.0117}.

\begin{equation}
\frac{\partial u}{\partial t} + c_x \frac{\partial u}{\partial x} + c_y \frac{\partial u}{\partial y} = 0, \quad c_x = 0.5, \quad c_y = 5.0
\end{equation}
\begin{equation}
u_0(x,y) = \exp\left[-60\left(x - 0.3y^2\right)^2\right] \sin\left(7(x + y)\right)
\end{equation}

\subsubsection{Anisotropic Ridge Advection}

In this example, we are covering a representation a atmospheric flow of a jet stream case, where the equation is:

\begin{equation}
\frac{\partial u}{\partial t} + c_x \frac{\partial u}{\partial x} + c_y \frac{\partial u}{\partial y} = 0, \quad c_x = 0.05, \quad c_y = 6.0
\end{equation}
\begin{equation}
u_0(x,y) = \exp\left[-60\left((x - y) - 1.0\right)^2\right] \sin\left(8(x + y)\right)
\end{equation}
This test models advection of a sharply defined ridge oriented at $45^\circ$ with extreme velocity anisotropy ($c_y/c_x = 120$). The solution develops strongly directional propagation with minimal diffusion, testing the model's ability to preserve oriented structures under biased transport.

\subsubsection{Sheared Kelvin-Helmholtz Stripes}
This is a models advection of sheared striped patterns with added Gaussian noise, resembling early-stage Kelvin-Helmholtz instability. Tests robustness to noise while preserving directional coherence \cite{fluids9030052}.

\begin{equation}
\frac{\partial u}{\partial t} + c_x \frac{\partial u}{\partial x} + c_y \frac{\partial u}{\partial y} = 0, \quad c_x = -2, \quad c_y = 0.5
\end{equation}
\begin{equation}
u_0(x,y) = \sin\left[12(x + 0.6y)\right] + 0.2\,\mathcal{N}(0,1)
\end{equation}

\subsubsection{Polygonal Shock Pattern}

This is a toy model where features a hexagonal shock pattern with sixfold symmetry. Challenges models with angular periodicity and discrete directional symmetries.

\begin{equation}
\frac{\partial u}{\partial t} + c_x \frac{\partial u}{\partial x} + c_y \frac{\partial u}{\partial y} = 0, \quad c_x = 2, \quad c_y = 1.3
\end{equation}
\begin{equation}
u_0(r,\theta) = \tanh\left[12\left(\cos(6\theta) - 0.3\right)\right]
\end{equation}

\subsection{Combined-effect problems.}

\subsubsection{Multi-Angle Shock Waves}

This is a 2D Burgers' equation with dual shock fronts propagating at angles $\theta_1 = \arctan(0.8)$ and $\theta_2 = \arctan(-1.2)$. Tests nonlinear interaction of directional discontinuities and shock-capturing capabilities. We intent to represent supersonic and combustion flows.

\begin{equation}
\frac{\partial u}{\partial t} + u \nabla u = \nu \nabla^2 u, \quad \nu = 8\times10^{-4}
\end{equation}
\begin{equation}
u_0(x,y) = \tanh\left[10(x + 0.8y - 3)\right] + \tanh\left[12(x - 1.2y - 1)\right]
\end{equation}

\subsubsection{Spiral Shock Pattern}
This is a geometrically complex test with an Archimedean spiral shock pattern. The orientation varies continuously with polar angle, requiring models to capture smoothly varying directional information.

\begin{equation}
\frac{\partial u}{\partial t} + u \nabla u = \nu \nabla^2 u, \quad \nu = 5\times10^{-4}
\end{equation}
\begin{equation}
u_0(r,\theta) = \tanh\left[10\left(r - 0.7 - 0.15\theta\right)\right]
\end{equation}

\section{Network Training}\label{sec:network_training}

The experiment trains and evaluates two neural operators on ten PDE simulation datasets. Both models take as input a 2D solution field $u_t \in \mathbb{R}^{N \times N}$ and predict the subsequent time step $u_{t+1}$. Thus, the learning task can be formulated as a \emph{supervised, single-step autoregressive regression} in the physical domain. The two architectures differ only in their spectral representations: FNO employs a \emph{global Fourier basis}, truncated to the lowest-frequency modes, whereas SNO uses a \emph{directional shearlet filterbank}, explicitly decomposing the field across both scale and orientation. Both models are trained using the configuration of the Table \ref{bhyper}.

\begin{table}[htbp]
\centering
\begin{tabular}{lll}
\toprule
\textbf{Parameter} & \textbf{Value} & \textbf{Notes} \\
\midrule
Channel width            & 8          & Hidden feature dimension \\
Learning rate            & $10^{-3}$  & AdamW \\
Weight decay             & $10^{-4}$  & Decoupled L2 regularization \\
Batch size               & 32         & Shuffled mini-batches \\
Maximum epochs           & 1000       & Subject to early stopping \\
Early stopping patience  & 500        & Epochs without improvement \\
Activation               & GELU       & After every layer \\
Loss function            & MSE        & $\|\hat{u}_{t+1} - u_{t+1}\|_2^2$ \\
Optimizer                & AdamW      & Adaptive + decoupled decay \\
\bottomrule
\end{tabular}
\caption{Both models' hyperparameters}
\label{bhyper}
\end{table}

More specifically, to ensure that SNO and FNO have approximately the same number of parameters, we selected the hyperparameter configurations for each model as summarized in Tables~\ref{fnotable} and~\ref{snotable}. This means a total of 12,000 parameters for SNO and 17,000 for FNO.

\begin{table}[htbp]
\centering
\begin{tabular}{lll}
\toprule
\textbf{Parameter} & \textbf{Value} & \textbf{Description} \\
\midrule
\texttt{modes x}            & 4    & Truncation in $k_x$ \\
\texttt{modes y}            & 4    & Truncation in $k_y$ \\
\texttt{spectral layers} & 4    & Number of FNO blocks \\
\bottomrule
\end{tabular}
\caption{Fourier Neural Operator hyperparameters}
\label{fnotable}
\end{table}

\begin{table}[htbp]
\centering
\begin{tabular}{lll}
\toprule
\textbf{Parameter} & \textbf{Value} & \textbf{Description} \\
\midrule
\texttt{n scales} ($J$)     & 4    & Dyadic radial bands \\
\texttt{n shears} ($S$)     & 64    & Angular directions per scale \\
\texttt{spectral layers} & 4    & Number of shearlet blocks \\
\bottomrule
\end{tabular}
\caption{Shearlet Neural Operator hyperparameters}
\label{snotable}
\end{table}

To evaluate the training performance of the shearlet and Fourier neural operators, the dataset is split chronologically to prevent temporal leakage. Specifically, 60\% of the data are used for training the architectures, 25\% for validation and early stopping, and the remaining 15\% for testing in a forward-in-time prediction setting.

Finally, in order to measure the quality of the training, the metrics below in Table \ref{metric} is used.

\begin{table}[htbp]
\centering
\begin{tabular}{lll}
\toprule
\textbf{Metric} & \textbf{Formula} & \textbf{Measures} \\
\midrule
MSE           & $\langle|\hat u - u|^2\rangle$         & Amplitude error \\
MAE           & $\langle|\hat u - u|\rangle$            & Robust to outliers \\
Relative L2   & $\sqrt{\mathrm{MSE}}\,/\,\|u\|_2$     & Scale-invariant \\
SSIM          & Structural Similarity Index            & Perceptual fidelity \\
\bottomrule
\end{tabular}
\caption{Metrics for architecture evaluation}
\label{metric}
\end{table}

The \textbf{relative L2 error} is the primary comparison metric. SSIM provides
complementary information about structural and orientation fidelity,
particularly relevant for the anisotropic PDE cases.


\section{Results and Discussions}\label{sec:results}
The primary goal, as mention before, is  to assess how the shearlet-based representation improves the learning of solution operators, particularly in scenarios characterized by anisotropic features, sharp gradients, and spatially localized discontinuities—conditions where conventional Fourier Neural Operators (FNOs) often struggle.

Our analysis first examines the predictive accuracy of SNO relative to FNO across seven PDE families encompassing nonlinear conservation laws. By directly comparing predictions, we quantify how the directional, multiscale, and spatially localized properties of shearlets contribute to capturing edges and shocks more faithfully than global Fourier modes.

In the following sections, we systematically present: (i) overall error metrics and accuracy gains, (ii) qualitative assessments of feature fidelity in regions with high anisotropy and discontinuities, and (iii) an analysis of the impact of shearlet-induced sparsity and directional sensitivity on the learning efficiency. Through this discussion, we highlight the contexts in which SNO offers the most significant improvements and provide insights into the architectural choices that enable these gains.
\subsection{Diffusion-dominated}
In this section, we present a diffusion-dominated case with a strongly anisotropic initial condition. The initial state consists of a superposition of multiple wave components with different initial phases, evolving in a medium characterized by anisotropic diffusion. This type of configuration is commonly encountered in biological systems. In particular, it mimics processes such as the spread of molecules in biological tissues, signal propagation along nerve fibers, and the transport of proteins through structured cellular environments, as discussed previously.
\subsection{Multi-Orientation Texture}
\begin{figure}[htbp]
    \centering
    \begin{subfigure}[b]{0.49\linewidth}
        \centering
        \includegraphics[width=\linewidth]{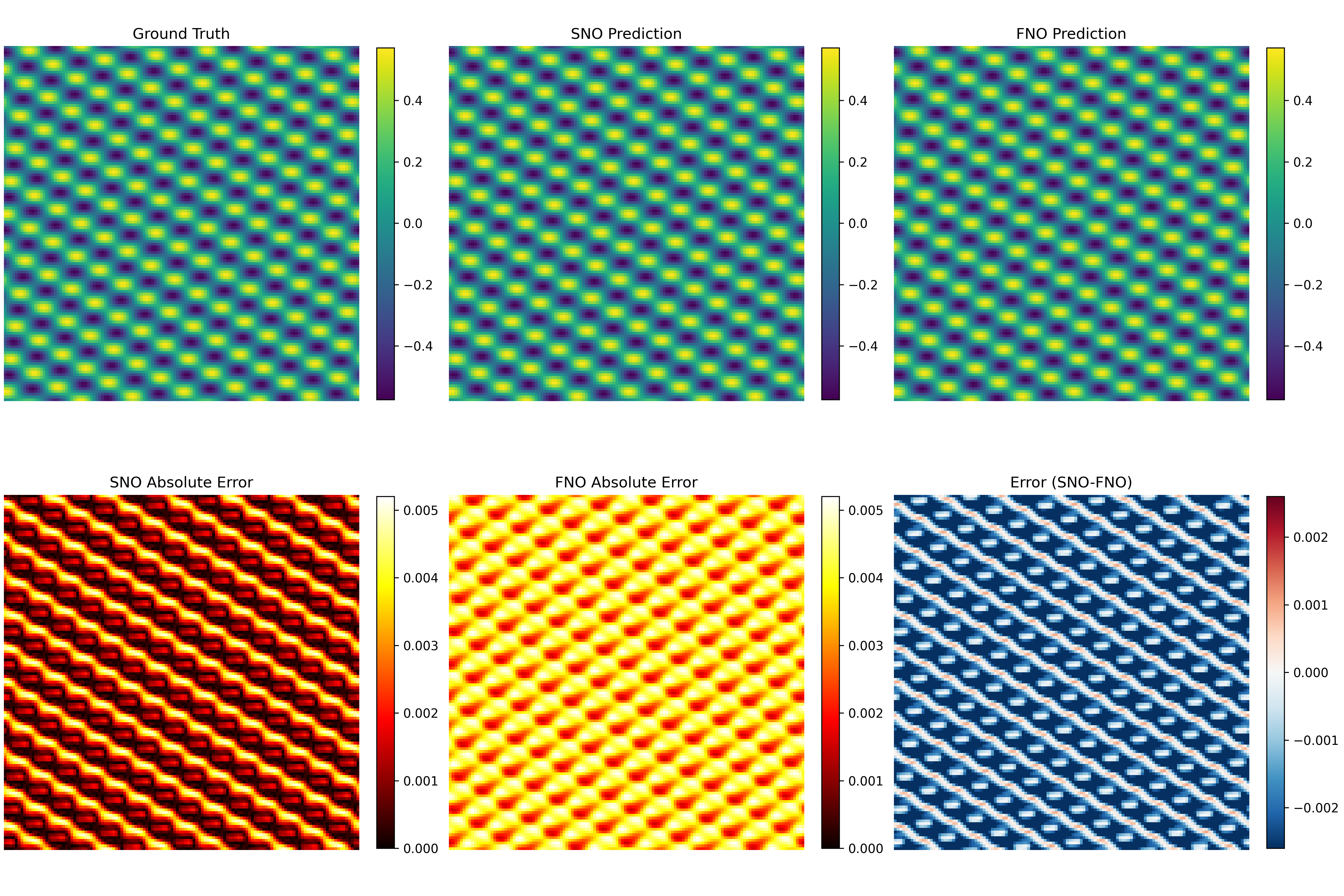}
        \caption{128$\times$128 mesh, T = 150}
        \label{fig:texture128}
    \end{subfigure}
    \hfill
    \begin{subfigure}[b]{0.49\linewidth}
        \centering
        \includegraphics[width=\linewidth]{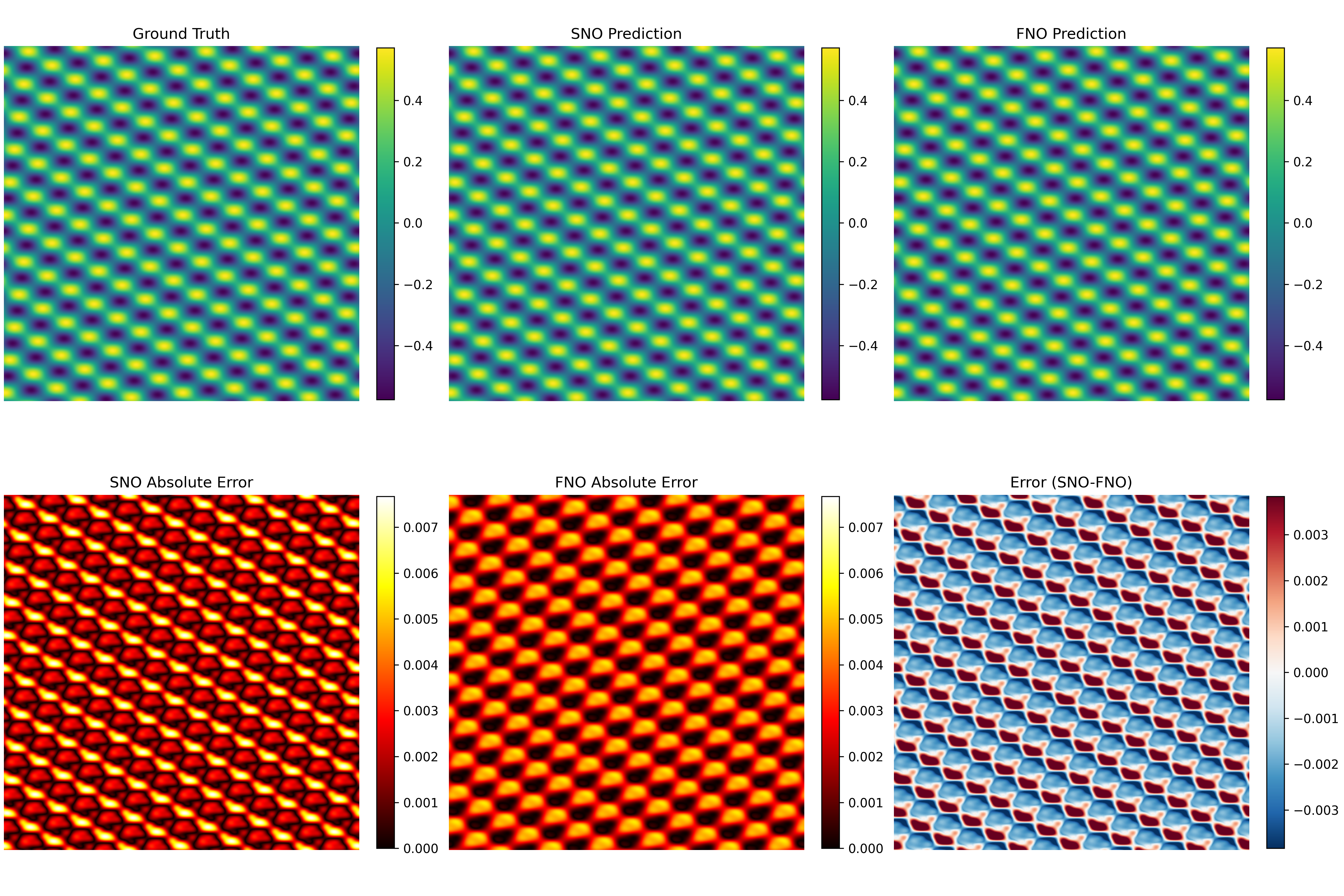}
        \caption{512$\times$512 mesh, T = 150}
        \label{fig:texture512}
    \end{subfigure}
    \caption{Comparison of the Ground Truth, SNO Prediction, FNO Prediction, their respective Absolute Errors, and the Error Difference between SNO and FNO for a diffusive dominant partial differential equation (PDE) solution.}
    \label{fig:maintexture}
\end{figure}

Figure \ref{fig:maintexture} illustrates the presence of multiple wave components exhibiting multiscale behavior, where both small- and large-scale waves interact and propagate through the medium at different rates.
One can observe in Figure~\ref{fig:maintexture} the presence of anisotropic patterns and layered structures in the solution. Both the SNO and FNO predictions closely resemble the Ground Truth, although small discrepancies are visible. However, the SNO exhibits smaller deviations compared to the FNO, indicating a more accurate reconstruction of the solution features. In particular, SNO more effectively captures the overall behavior of the field, including oscillatory and multi-textured structures present in the data. While the FNO prediction preserves the main structure of the solution, it appears less precise, with more noticeable deviations from the Ground Truth than those observed in the SNO result.

As can be observed FNO Absolute Error shows larger discrepancies compared to the SNO prediction, since SNO better preserves high-frequency directional textures. while FNO introduces smoothing and phase errors in oriented patterns.

The Error Difference between SNO and FNO highlights the discrepancies between the two models' predictions. The difference is small in most areas, but we see some regions where FNO has larger errors than SNO. These regions are likely where FNO is struggling to capture the solution's finer features, as evidenced by the higher error magnitudes in the FNO Absolute Error panel. Also from Figure~\ref{fig:maintexture}, the differences appear as smooth oscillations, and the SNO model appears to outperform FNO in these regions. These errors are likely to occur because FNO's Fourier basis struggles to capture the multi-scale or texture-like features present in the solution, which is often the case for non-smooth solutions like those found in diffusion-dominated problems. This can be observed either with N = 128 or N = 512. However, from Figure~\ref{fig:maintexture}, we can see that with increase the mesh size, the dataset, FNO increase its accurancy. It means that SNO needs less data.

\begin{figure}[htbp]
    \centering
    \begin{subfigure}[b]{0.49\linewidth}
        \centering
    \includegraphics[width=\textwidth]{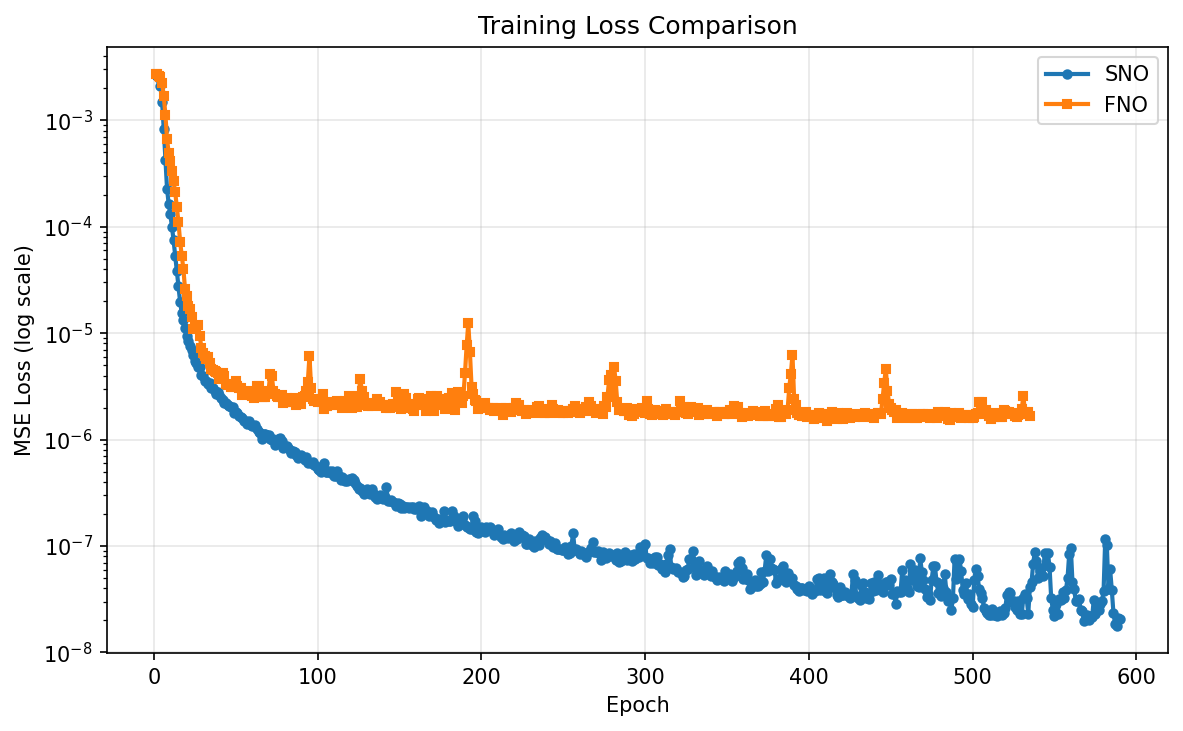}
        \caption{Training with 128$\times$128 mesh size for diffusive multi-orientation texture case.}
        \label{fig:sub1texture}
    \end{subfigure}
    \hfill 
    \begin{subfigure}[b]{0.49\linewidth}
        \centering
    \includegraphics[width=\textwidth]{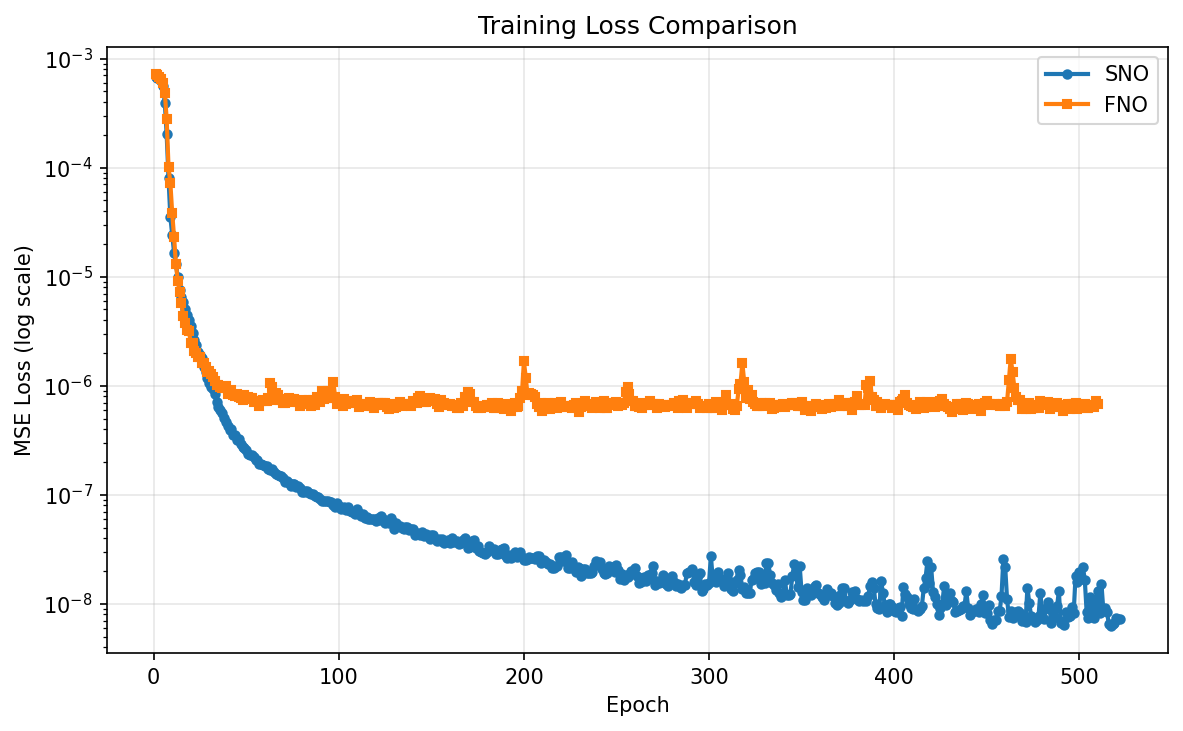}
        \caption{Training with 512$\times$512 mesh size for diffusive multi-orientation texture case.}
        \label{fig:lossTexture}
    \end{subfigure}
    \caption{Training loss comparison between SNO and FNO. The loss (MSE) is plotted on a logarithmic scale against the number of epochs for diffusive multi-orientation texture case.}
    \label{fig:mainlosstexture}
\end{figure}

In summary, both operators approximate the PDE solution well at moderate resolution. However, the absolute error of FNO exhibits larger discrepancies compared to the SNO prediction. This indicates that SNO is more effective at learning and capturing the underlying structures of diffusion-dominated PDEs when limited grid resolution is available. In particular, due to its multi-scale and directional representation, SNO is better suited to represent multi-textured and anisotropic features, enabling more accurate approximations with fewer grid points and less training data.
Figure \ref{fig:mainlosstexture} presents a comparative analysis of the training dynamics between the Spectral Neural Operator (SNO) and the Fourier Neural Operator (FNO) by plotting the Mean Squared Error (MSE) loss on a logarithmic scale as a function of training epochs. The results show that SNO achieves lower error levels even when trained with lower-resolution data (N = 128), whereas FNO requires significantly higher resolution. 
Figure \ref{fig:mainlosstexture} presents a comparative analysis of the training dynamics between the Spectral Neural Operator (SNO) and the Fourier Neural Operator (FNO) by plotting the Mean Squared Error (MSE) loss on a logarithmic scale as a function of training epochs. The results show that SNO achieves lower error levels even when trained with lower-resolution data (N = 128), whereas FNO requires significantly higher resolution N=512) to reach comparable error reduction.
This indicates that SNO is able to more efficiently represent the dynamics of this diffusion-dominated case characterized by strong anisotropy and multiple spatial scales, even with a smaller dataset. Furthermore, the training curves reveal that the FNO loss quickly reaches a plateau, while SNO continues to decrease its MSE throughout training. This behavior suggests that SNO retains additional capacity to further improve the solution approximation, whereas FNO appears to saturate earlier in the optimization process.

\subsection{Convection-dominated}

For hyperbolic equations, the Fourier Neural Operator (FNO) may exhibit limitations when discontinuities are present. To assess its performance under such conditions, we consider a convection-dominated scenario. In particular, we analyze three cases in which heterogeneous and anisotropic initial conditions are transported over time.

\subsubsection{Bent and Anisotropic Ridge Advection, Sheared Kelvin-Helmholtz Stripes and Polygonal Shock Pattern}

\begin{figure}[htbp]
    \centering
    \begin{subfigure}[b]{0.49\linewidth}
        \centering
    \includegraphics[width=\textwidth]{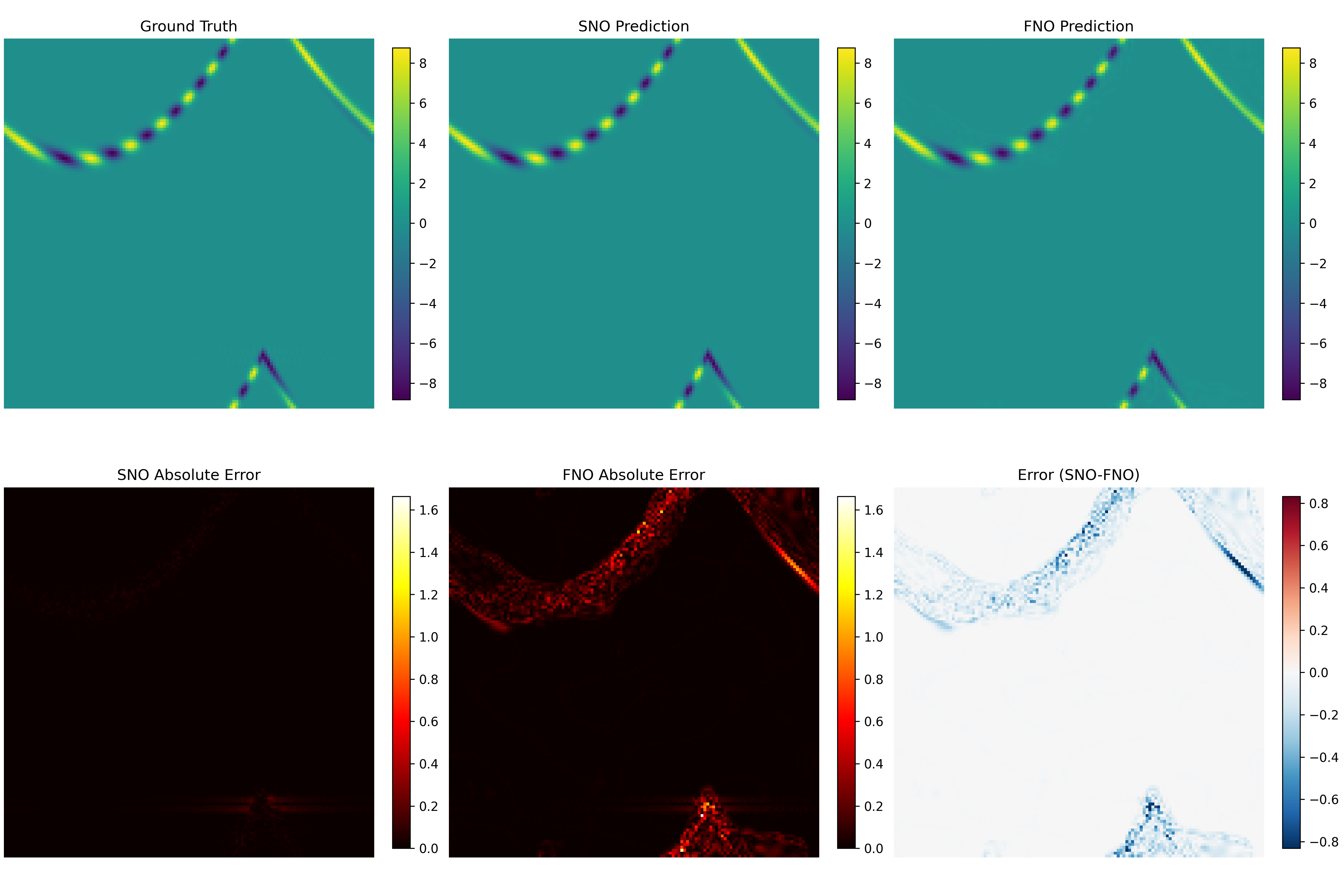}
        \caption{Solution with 128$\times$128 mesh size for bent ridge advect case in T = 150.}
        \label{fig:ridge}
    \end{subfigure}
    \hfill 
    \begin{subfigure}[b]{.49\linewidth}
        \centering
    \includegraphics[width=\textwidth]{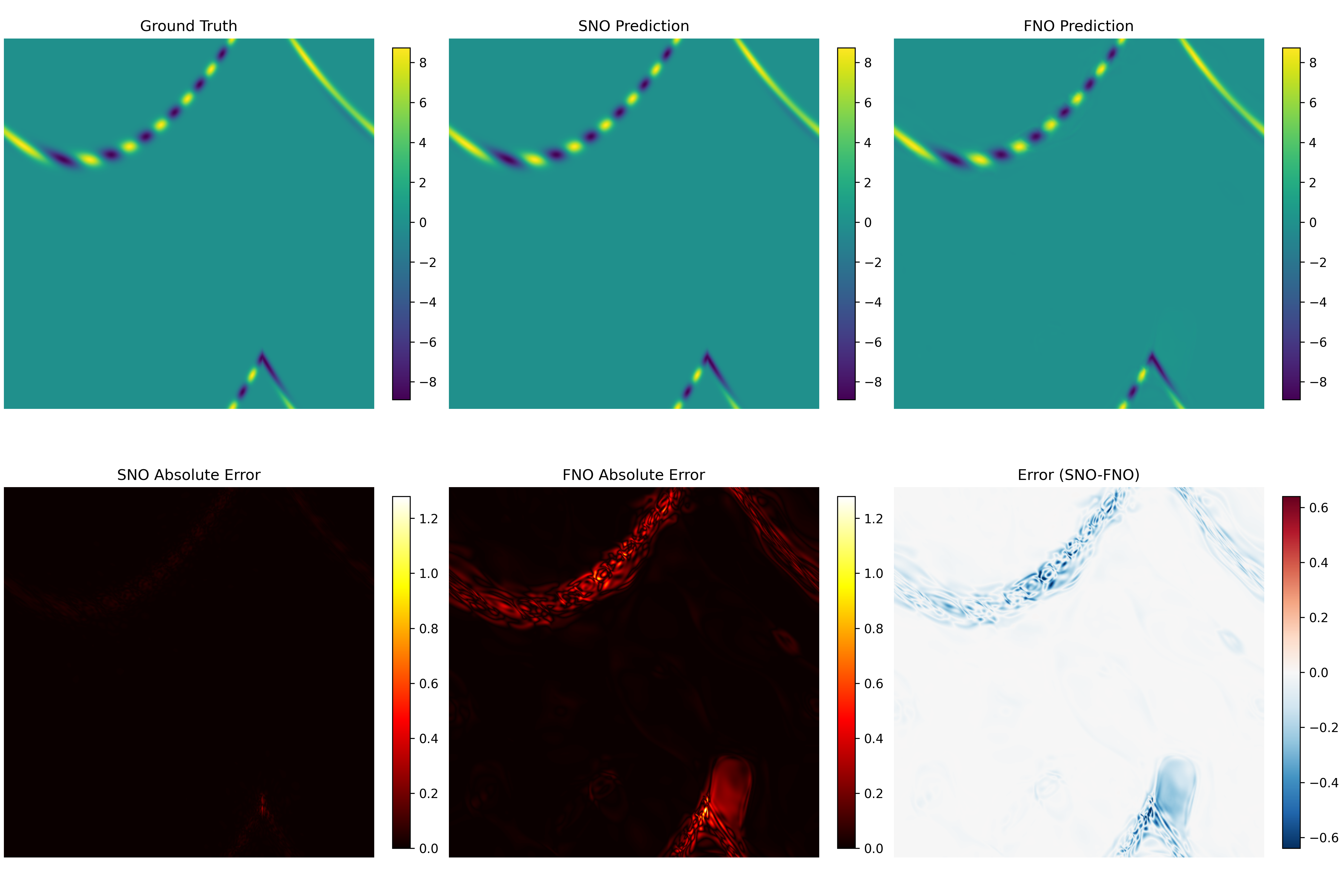}
        \caption{Solution with 512$\times$512 mesh size for bent ridge advect case in T = 150.}
        \label{fig:ridge512}
    \end{subfigure}
    \caption{Comparison of the Ground Truth, SNO Prediction, FNO Prediction, their respective Absolute Errors, and the Error Difference between SNO and FNO for bent ridge advvect partial differential equation (PDE) solution.}
    \label{fig:mainBent}
\end{figure}

\begin{figure}[htbp]
    \centering
    \begin{subfigure}[b]{.49\linewidth}
        \centering
    \includegraphics[width=\textwidth]{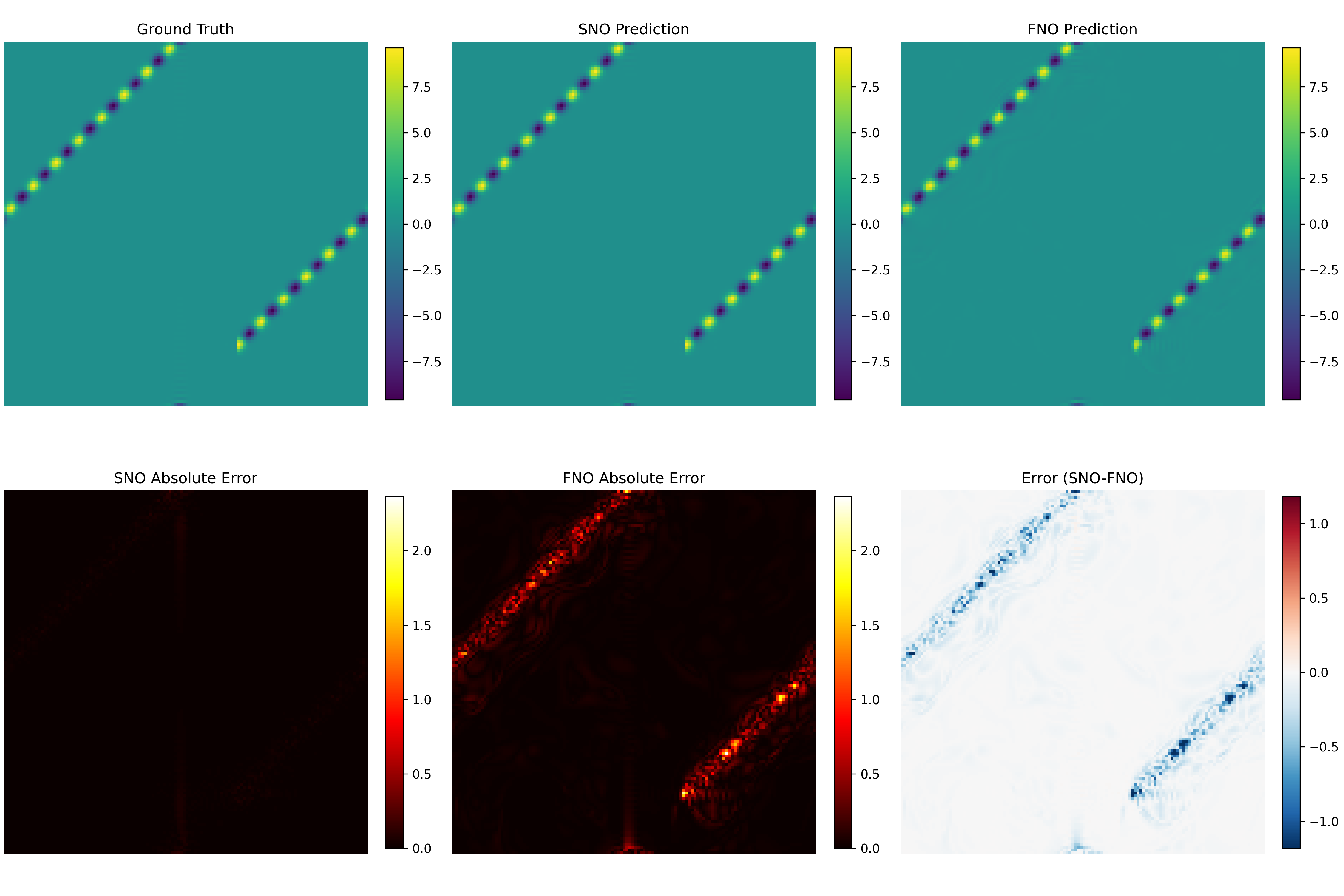}
        \caption{Solution with 128$\times$128 mesh size for anisotropic ridge advect case in T = 150.}
        \label{fig:aridge}
    \end{subfigure}
    \hfill 
    \begin{subfigure}[b]{0.49\linewidth}
        \centering
    \includegraphics[width=\textwidth]{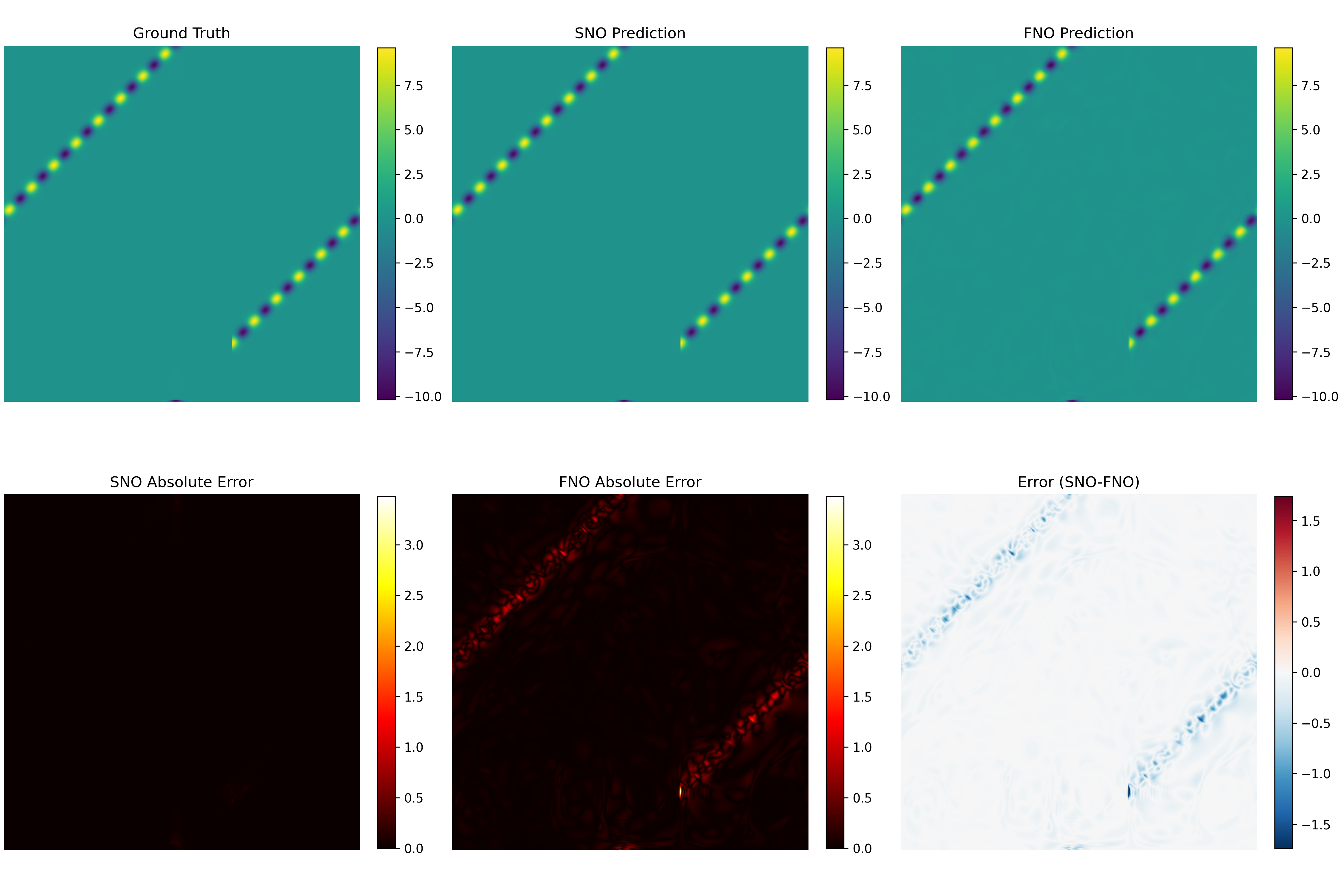}
        \caption{Solution with 512$\times$512 mesh size for anisotropic ridge advect case in T = 150.}
        \label{fig:aridge512}
    \end{subfigure}
    \caption{Comparison of the Ground Truth, SNO Prediction, FNO Prediction, their respective Absolute Errors, and the Error Difference between SNO and FNO for an anisotropic ridge advect differential equation (PDE) solution.}
    \label{fig:mainaniso}
\end{figure}

\begin{figure}[htbp]
    \centering
    \begin{subfigure}[b]{0.49\linewidth}
        \centering
    \includegraphics[width=\textwidth]{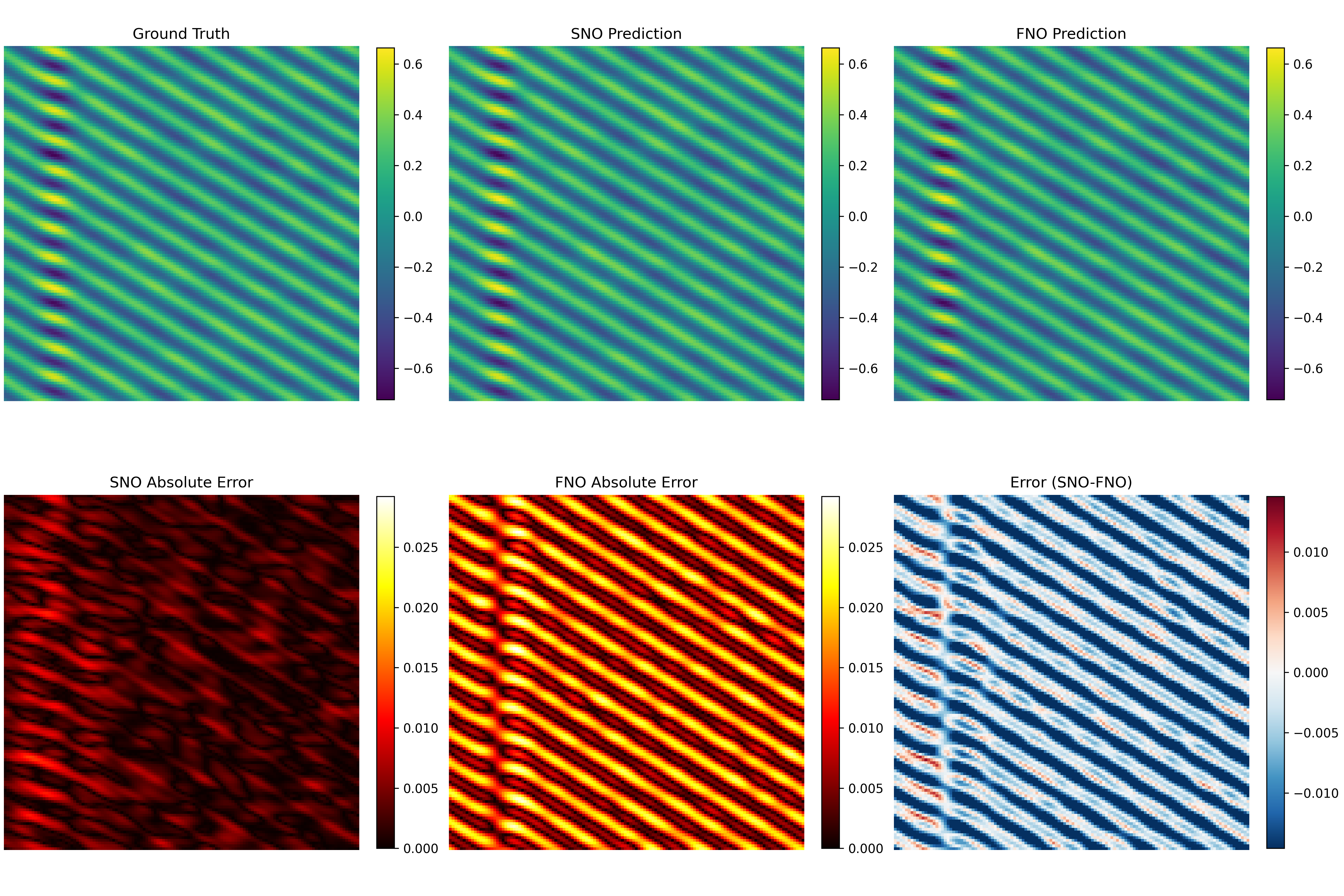}
        \caption{Solution with 128$\times$128 mesh size for sheared Kelvin-Helmholtz Stripes case in T = 100.}
        \label{fig:kelvin}
    \end{subfigure}
    \hfill 
    \begin{subfigure}[b]{0.49\linewidth}
        \centering
    \includegraphics[width=\textwidth]{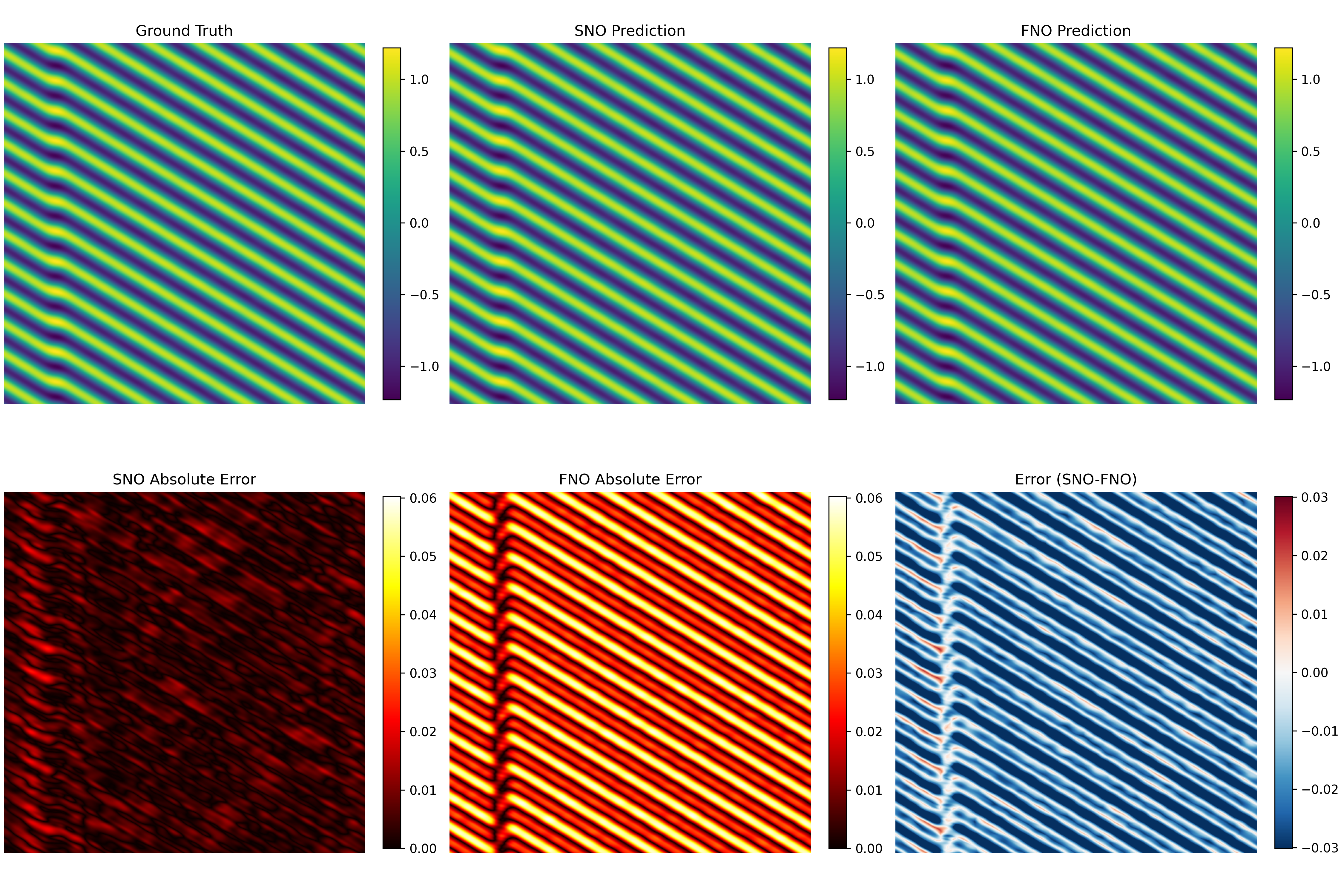}
        \caption{Solution with 512$\times$512 mesh size for sheared Kelvin-Helmholtz Stripes case in T = 100.}
        \label{fig:kelvin512}
    \end{subfigure}
    \caption{Comparison of the Ground Truth, SNO Prediction, FNO Prediction, their respective Absolute Errors, and the Error Difference between SNO and FNO for the sheared Kelvin-Helmholtz Stripes partial differential equation (PDE) solution}
    \label{fig:mainsheared}
\end{figure}

\begin{figure}[htbp]
    \centering
    \begin{subfigure}[b]{0.49\linewidth}
        \centering
    \includegraphics[width=\textwidth]{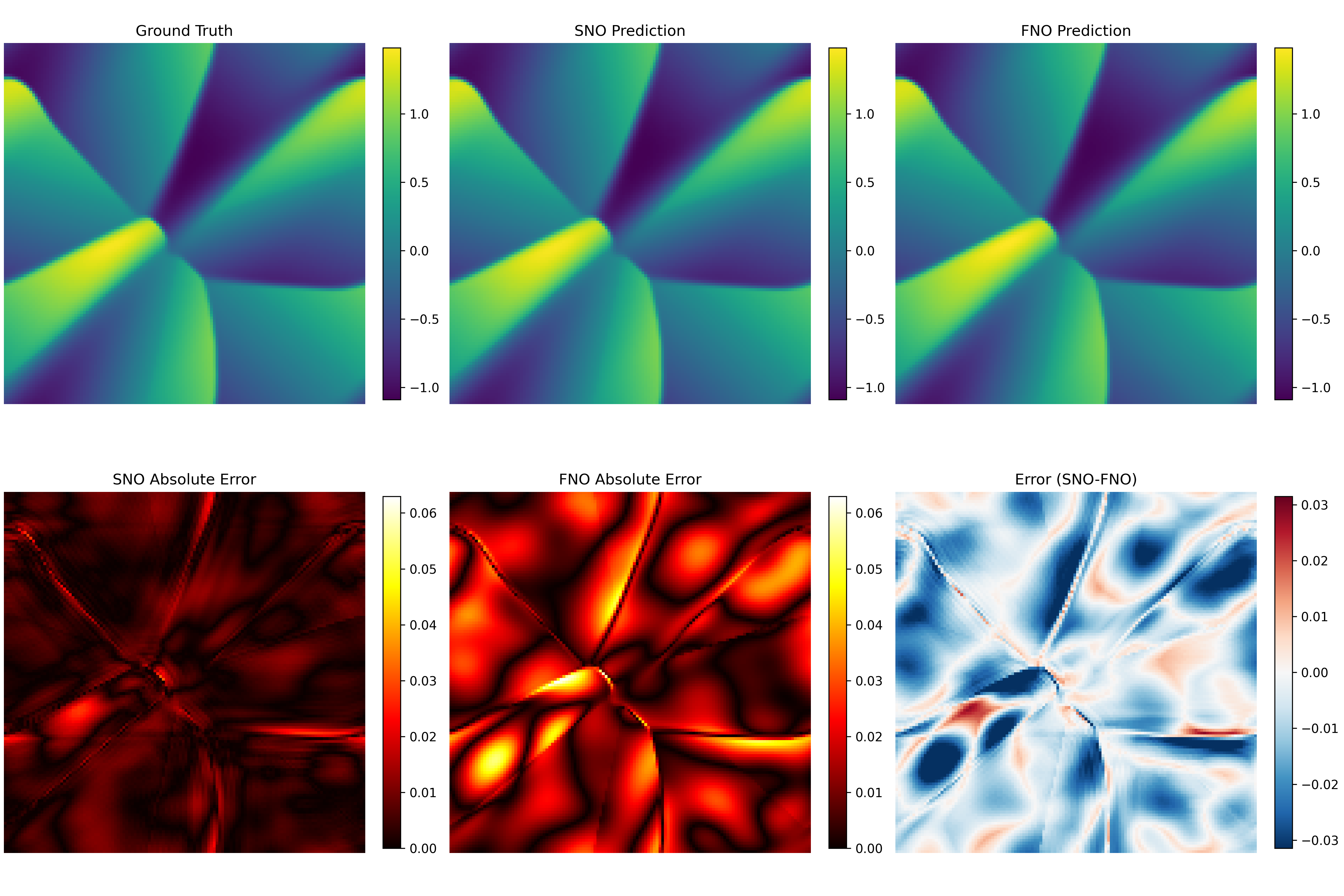}
        \caption{Solution with 128$\times$128 mesh size for polygonal shock case in T = 220.}
        \label{fig:polygonal}
    \end{subfigure}
    \hfill 
    \begin{subfigure}[b]{0.49\linewidth}
        \centering
    \includegraphics[width=\textwidth]{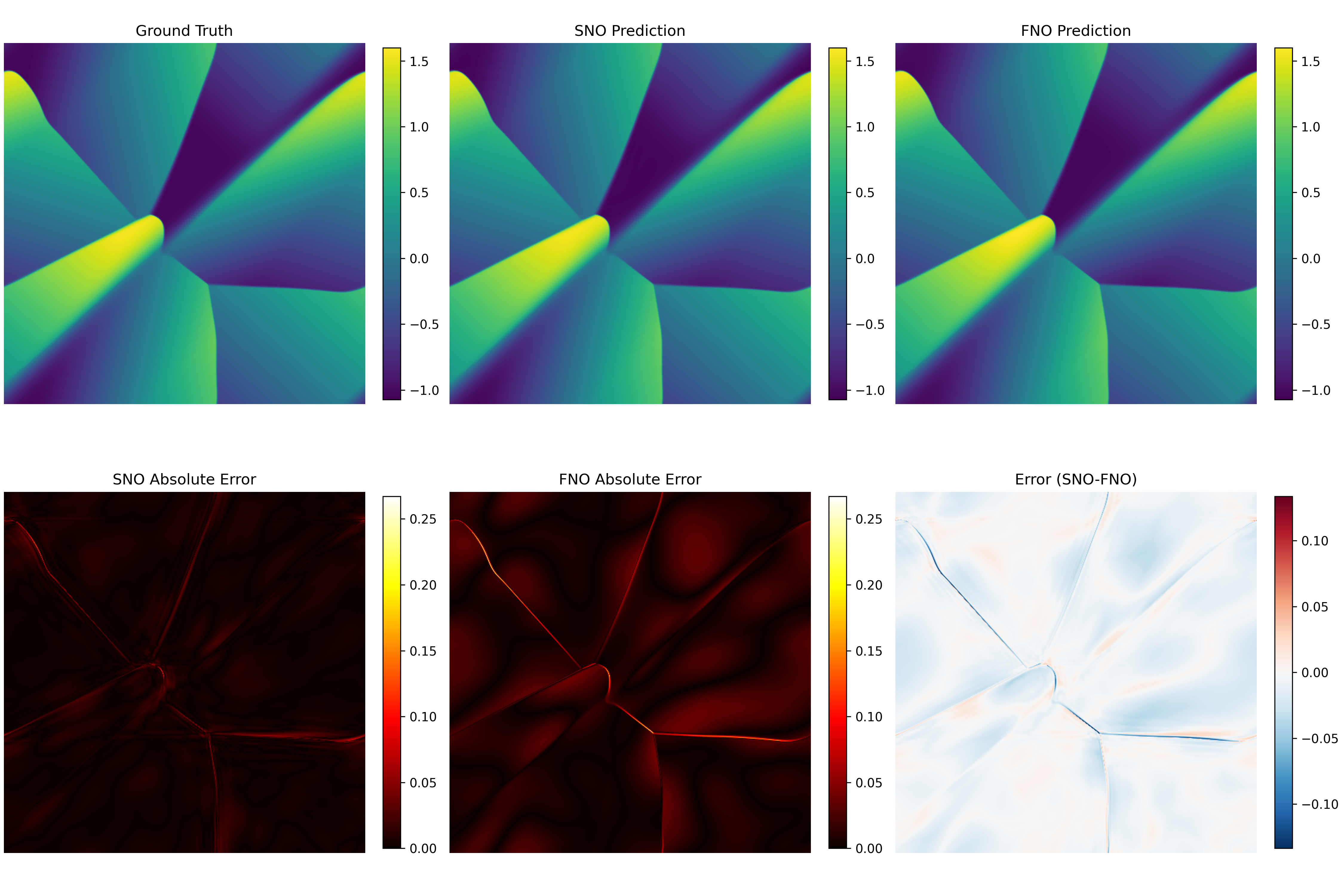}
        \caption{Solution with 512$\times$512 mesh size for polygonal shock case in T = 220.}
        \label{fig:polygonal512}
    \end{subfigure}
    \caption{Comparison of the Ground Truth, SNO Prediction, FNO Prediction, their respective Absolute Errors, and the Error Difference between SNO and FNO for a polygonal shock case partial differential equation (PDE) solution.}
    \label{fig:mainpolygonal}
\end{figure}

For convection-dominated PDEs, solutions often exhibit propagating fronts, shocks, filaments, and anisotropic wave packets—structures that are sparsely representable in shearlet systems due to their optimality for cartoon-like images with edges. This is evident in Figures~\ref{fig:mainBent},~\ref{fig:mainaniso},~\ref{fig:mainsheared} and~\ref{fig:mainpolygonal}, which illustrate propagating wave packets and anisotropic instabilities that are highly localized and directional in nature.

The advantage of the Shearlet Neural Operator (SNO) stems from its use of a representation basis that is inherently directional and local. In contrast, the Fourier Neural Operator (FNO) relies on a global, non-localized, and isotropic basis. Shearlets are specifically designed to capture anisotropic structures and directional features through a combination of scales, shears (encoding direction), and translations. This enables them to represent diagonal and curvilinear features—such as those found in transport-dominated flows—very sparsely.

This interpretation is supported by the SNO error map, which exhibits near-zero error throughout the domain, with only faint residual structures near wave centers. Moreover, the differences in error between SNO and FNO are strongly localized along transport trajectories, precisely where gradients are largest. These observations strongly suggest that the performance gain of SNO emerges in regions where directional transport dominates.

\begin{figure}[htbp]
    \centering
    \begin{subfigure}[b]{0.49\linewidth}
        \centering
    \includegraphics[width=\textwidth]{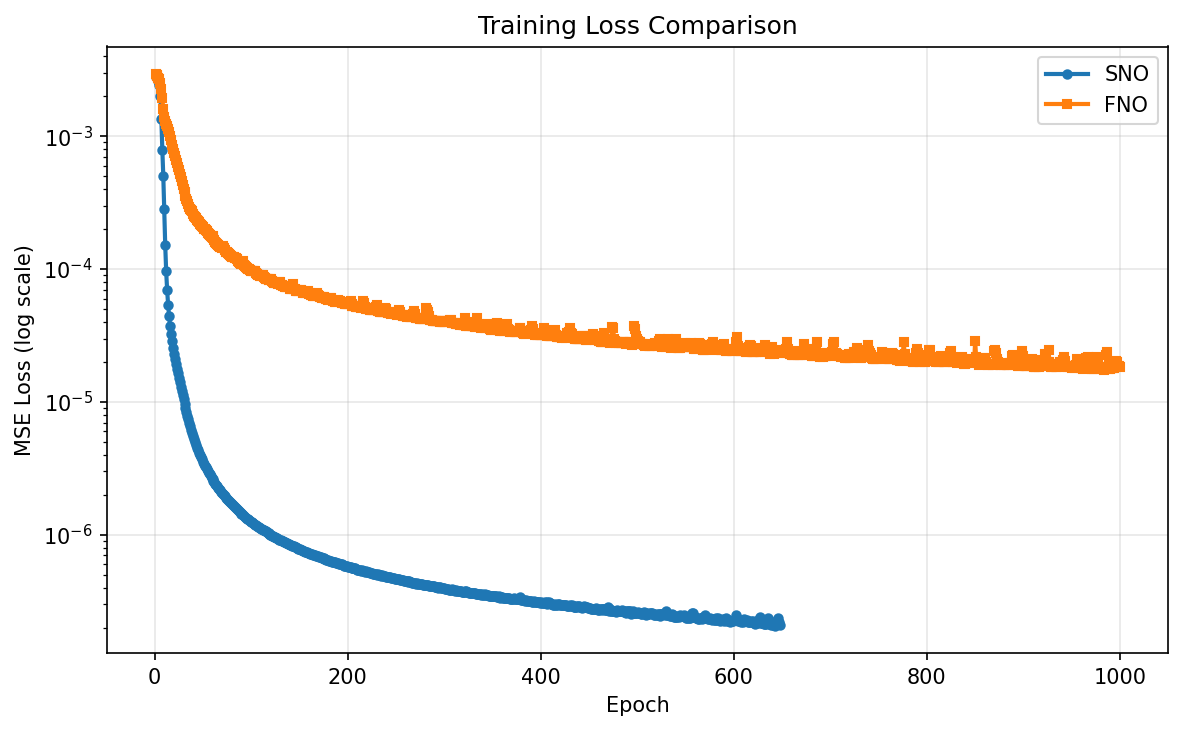}
        \caption{Training loss for mesh 128$\times$128 mesh size for bent ridge advect.}
        \label{fig:sub1bent}
    \end{subfigure}
    \hfill 
    \begin{subfigure}[b]{0.49\linewidth}
        \centering
    \includegraphics[width=\textwidth]{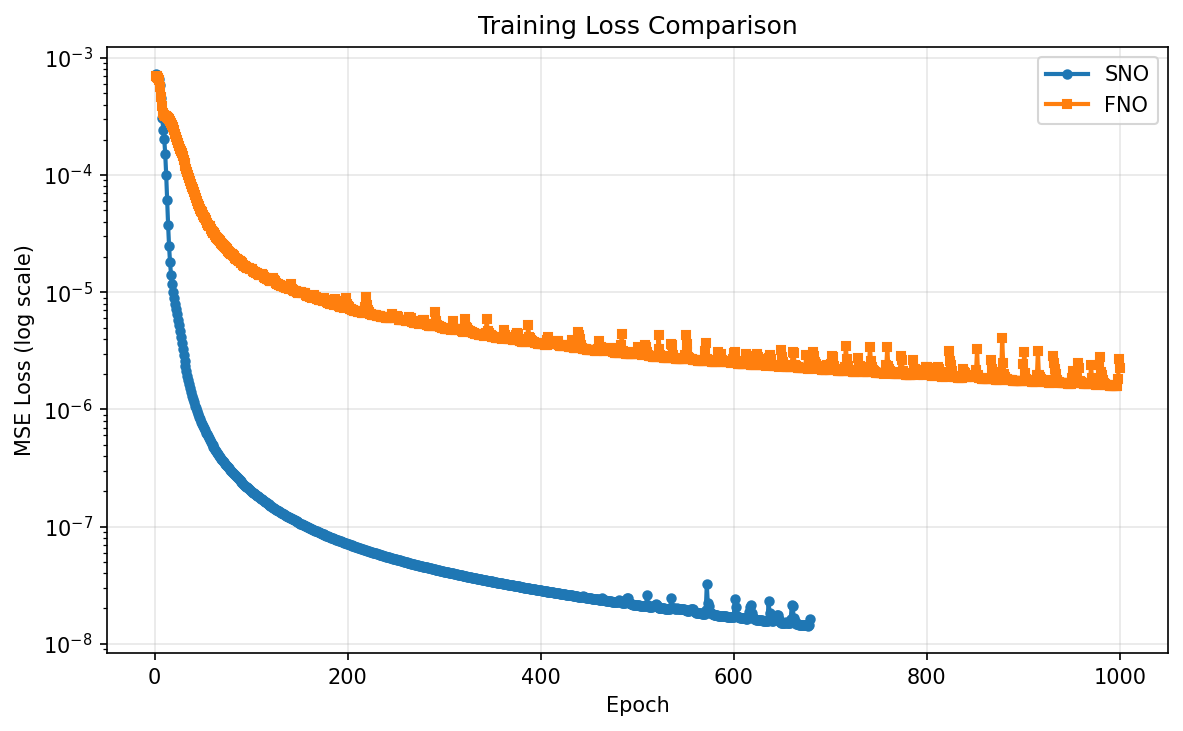}
        \caption{Training loss for mesh 512$\times$512 mesh size for bent ridge advect}
        \label{fig:sub2bent}
    \end{subfigure}
    \caption{Training loss comparison for bent ridge advect between SNO and FNO. The loss (MSE) is plotted on a logarithmic scale against the number of epochs for bent ridge advect case.}
    \label{fig:mainlossBent}
\end{figure}

\begin{figure}[htbp]
    \centering
    \begin{subfigure}[b]{0.49\linewidth}
        \centering
    \includegraphics[width=\textwidth]{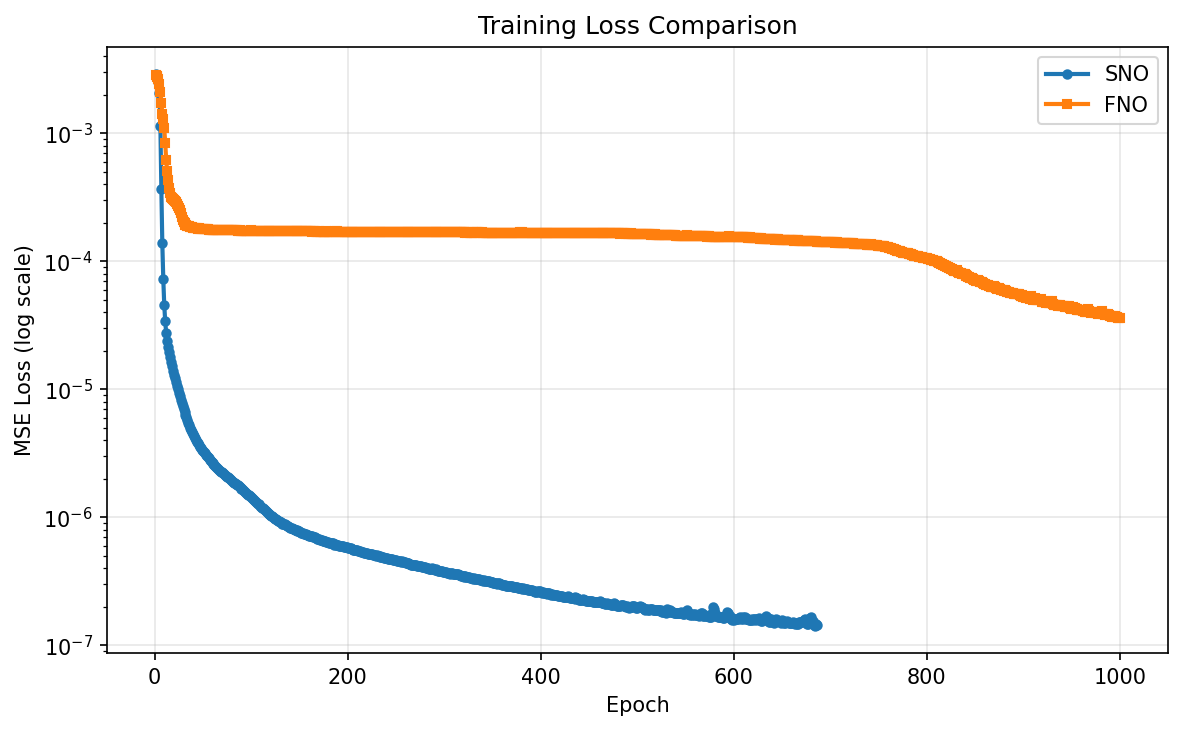}
        \caption{Training loss for mesh 128$\times$128 mesh size for anisotropic ridge advect}
        \label{fig:sub1ridge}
    \end{subfigure}
    \hfill 
    \begin{subfigure}[b]{0.49\linewidth}
        \centering
    \includegraphics[width=\textwidth]{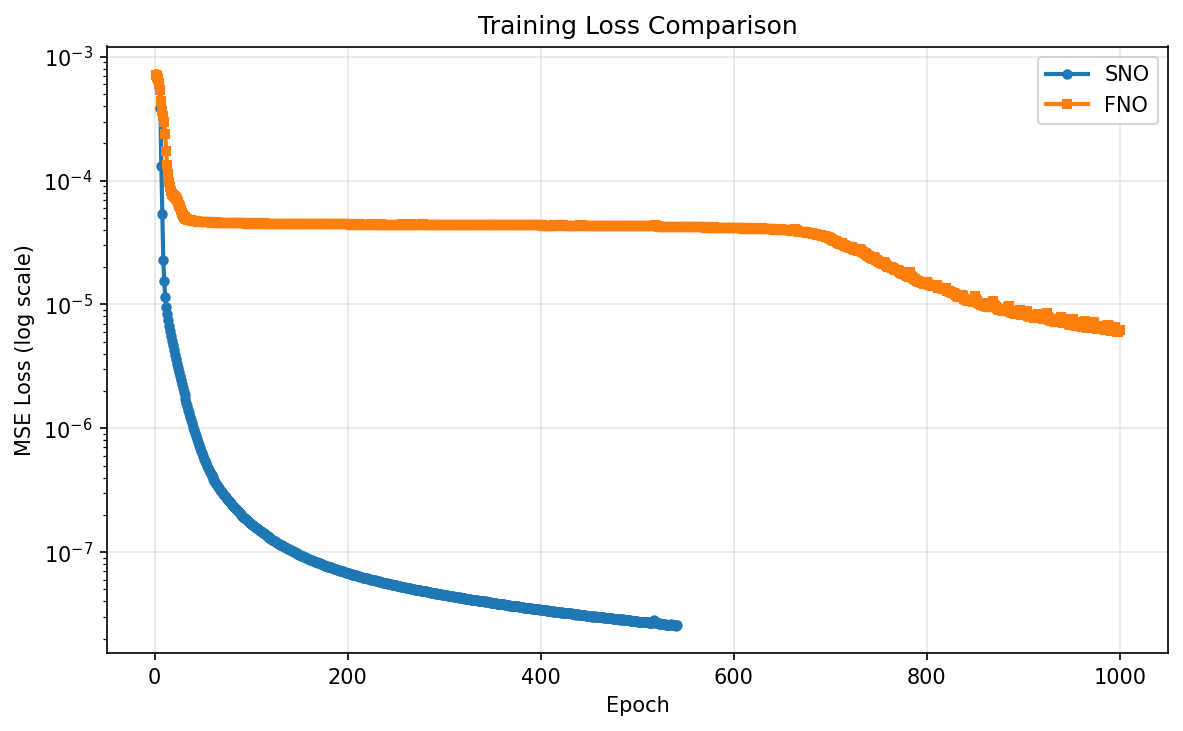}
        \caption{Training loss for mesh 512$\times$512 mesh size for anisotropic ridge advect}
        \label{fig:sub2ridge}
    \end{subfigure}
    \caption{Training loss comparison for anisotropic ridge advect between SNO and FNO. The loss (MSE) is plotted on a logarithmic scale against the number of epochs for anisotropic ridge advect case.}
    \label{fig:mainlossaniso}
\end{figure}

\begin{figure}[htbp]
    \centering
    \begin{subfigure}[b]{0.49\linewidth}
        \centering
    \includegraphics[width=\textwidth]{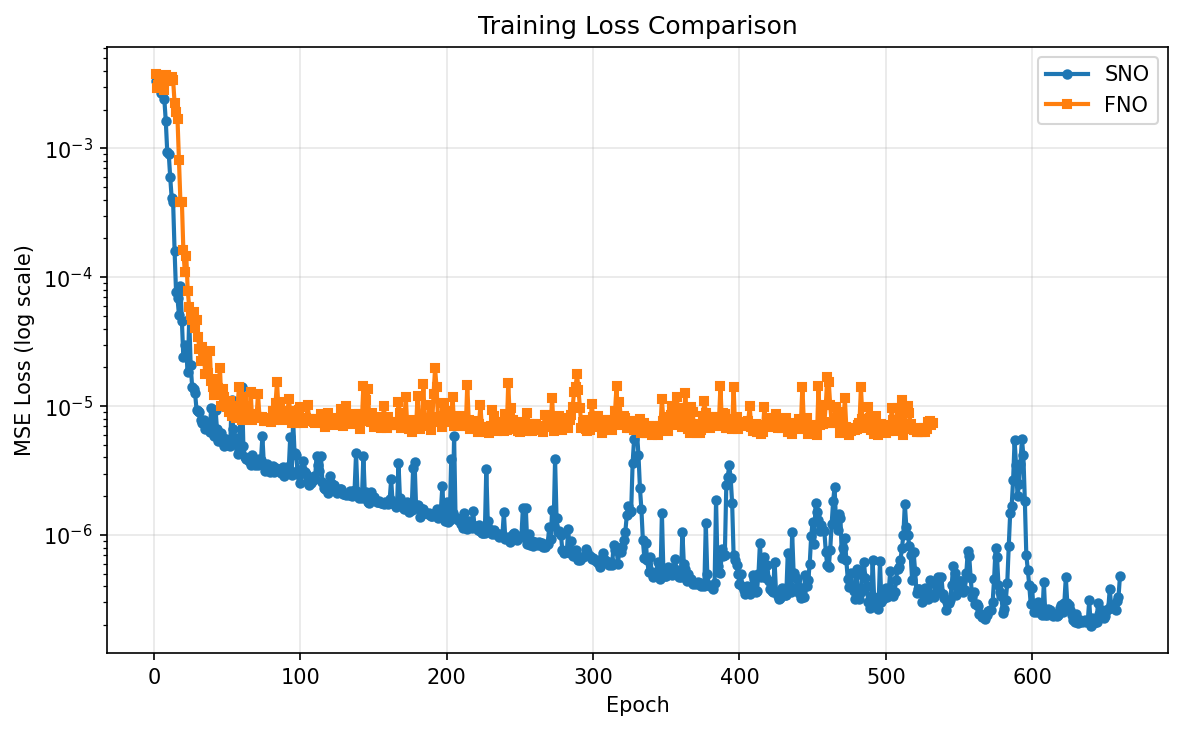}
        \caption{Training loss for mesh 128$\times$128 mesh size for sheared Kelvin-Helmholtz stripes case.}
        \label{fig:sub1kelvin}
    \end{subfigure}
    \hfill 
    \begin{subfigure}[b]{0.49\linewidth}
        \centering
    \includegraphics[width=\textwidth]{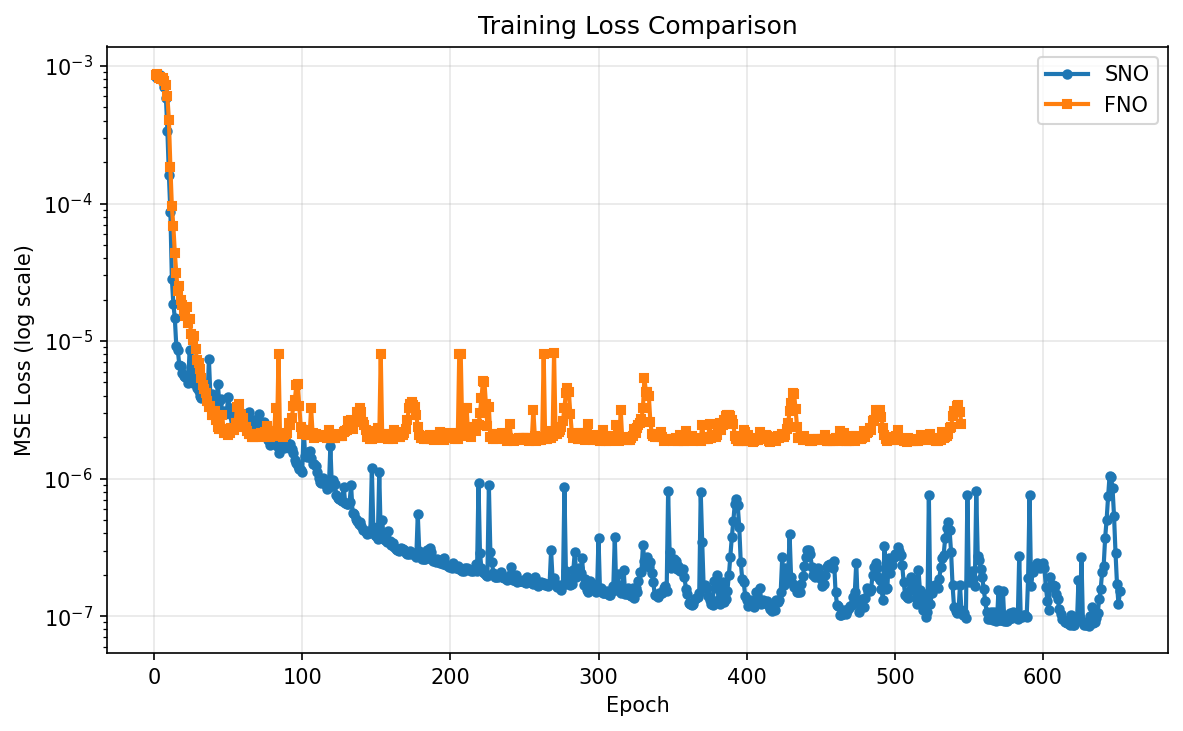}
        \caption{Training loss for mesh 512$\times$512 mesh size for sheared Kelvin-Helmholtz stripes case.}
        \label{fig:sub2kelvin}
    \end{subfigure}
    \caption{Training loss comparison for sheared Kelvin-Helmholtz Stripes between SNO and FNO. The loss (MSE) is plotted on a logarithmic scale against the number of epochs for sheared Kelvin-Helmholtz stripes case.}
    \label{fig:mainlosssheared}
\end{figure}

Figures \ref{fig:mainlosssheared}, \ref{fig:mainlossaniso}, and \ref{fig:mainlossBent} provide a systematic comparison of the training dynamics between two prominent neural operator architectures: the Spectral Neural Operator (SNO) and the Fourier Neural Operator (FNO). This analysis investigates the convergence behavior of both models by tracking the Mean Squared Error (MSE) loss on a logarithmic scale throughout the training process.

\begin{figure}[htbp]
    \centering
    \begin{subfigure}[b]{0.49\linewidth}
        \centering
        \includegraphics[width=\textwidth]{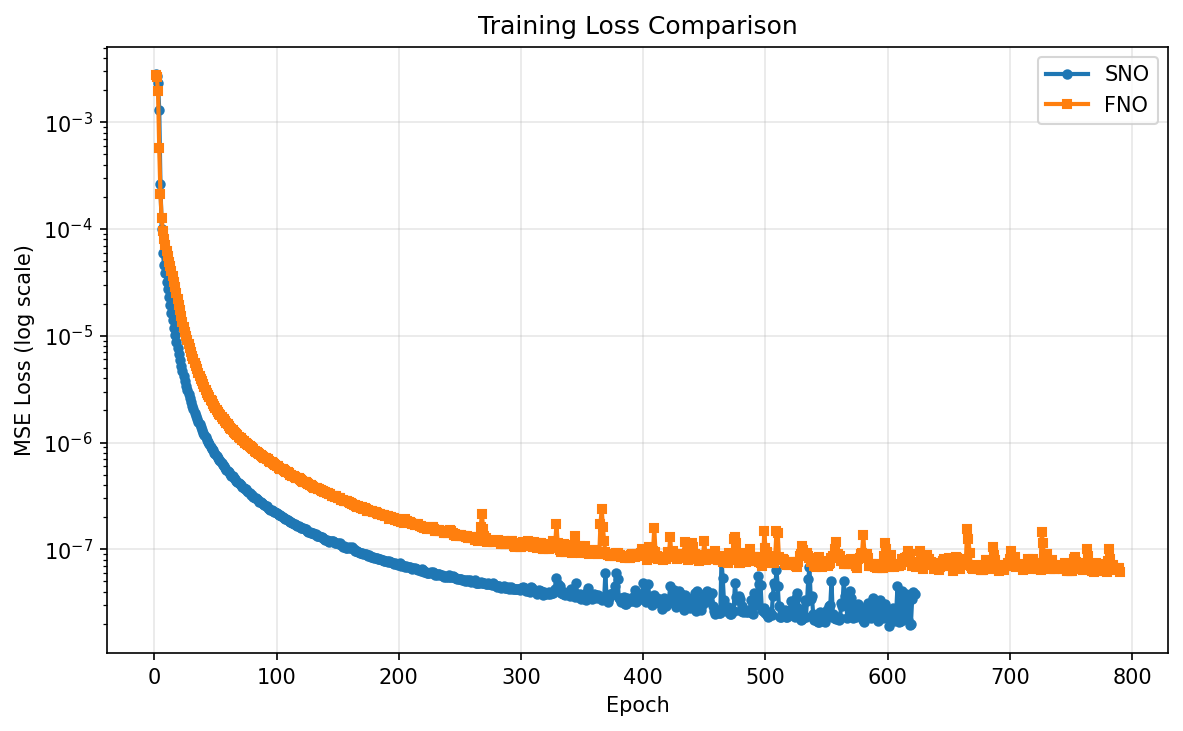}
        \caption{Training loss for mesh $128 \times 128$ mesh size for polygonal\_shock case.}
        \label{fig:sub1poly}
    \end{subfigure}
    \hfill
    \begin{subfigure}[b]{0.49\linewidth}
        \centering
        \includegraphics[width=\textwidth]{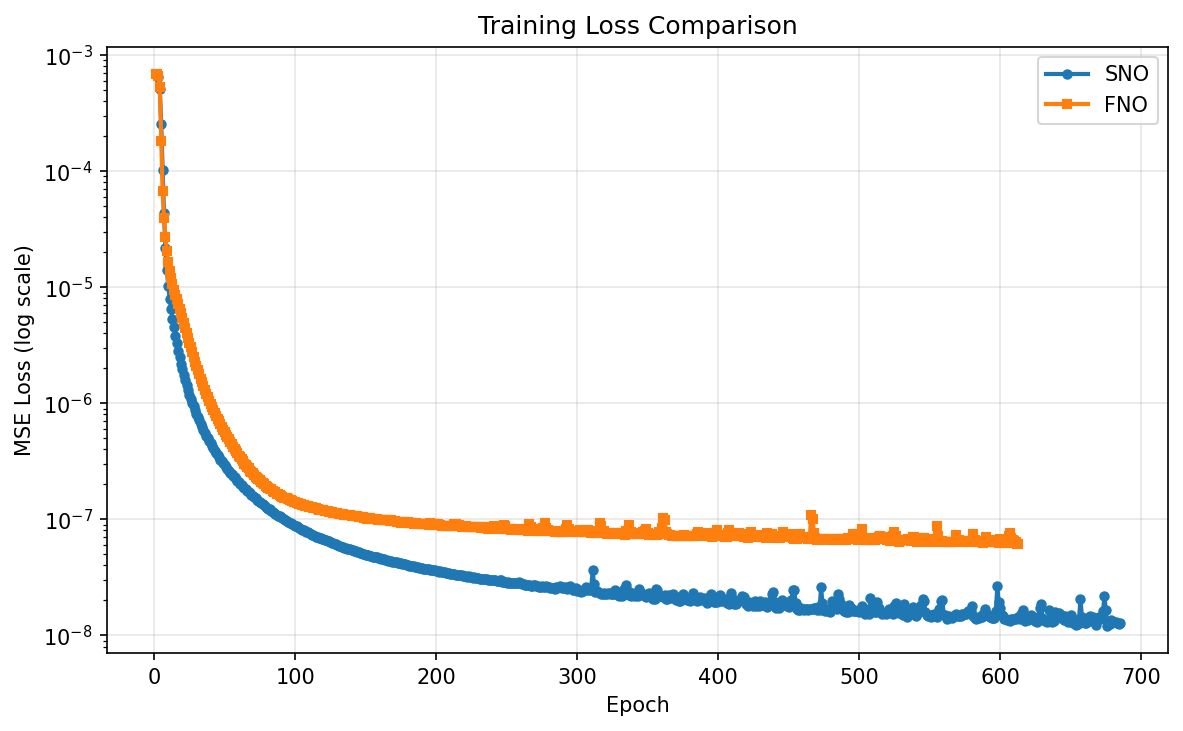}
        \caption{Training loss for mesh $512 \times 512$ mesh size for polygonal shock case.}
        \label{fig:sub2poly}
    \end{subfigure}
    \caption{Training loss comparison for polygonal shock case between SNO and FNO. The loss (MSE) is plotted on a logarithmic scale against the number of epochs for polygonal shock.}
    \label{fig:mainlosspolygonal}
\end{figure}

From the loss plots, we observe that the SNO converges rapidly to a value near $10^{-8}$, while the FNO plateaus at approximately $10^{-5}$. This indicates that the final error of the FNO is roughly two or three orders of magnitude higher than that of the SNO. However, as shown in Figure \ref{fig:mainlosspolygonal}, this gap narrows when the dataset is modified, though the FNO error still remains about one order of magnitude larger. In both cases, increasing the mesh resolution—and thus the dataset size—improves the accuracy of both neural operators, consistently reducing the error and mean squared error (MSE). These results suggest that the SNO is particularly well-suited for advection-dominated problems involving propagating discontinuities, such as those encountered in multiphase flow dynamics and other natural phenomena.

\subsection{Combined-effect problems.}

This combined effect exemplifies a low-viscosity regime, wherein solutions develop discontinuities—specifically shock formations—that pose significant challenges for accurate emulation by neural operators. The objective of neural operators, such as the Fourier Neural Operator (FNO) and Spectral Neural Operator (SNO), is to learn the solution operator that maps initial conditions to the corresponding solution fields at future time points. In this work, we analyze two distinct cases of the Burgers equation, each characterized by different initial conditions designed to emulate diverse physical scenarios.

\subsubsection{Multi-Angle Shock Waves and Spiral Shock Pattern}

\begin{figure}[htbp]
    \centering
    \begin{subfigure}[b]{0.49\linewidth}
        \centering
    \includegraphics[width=\textwidth]{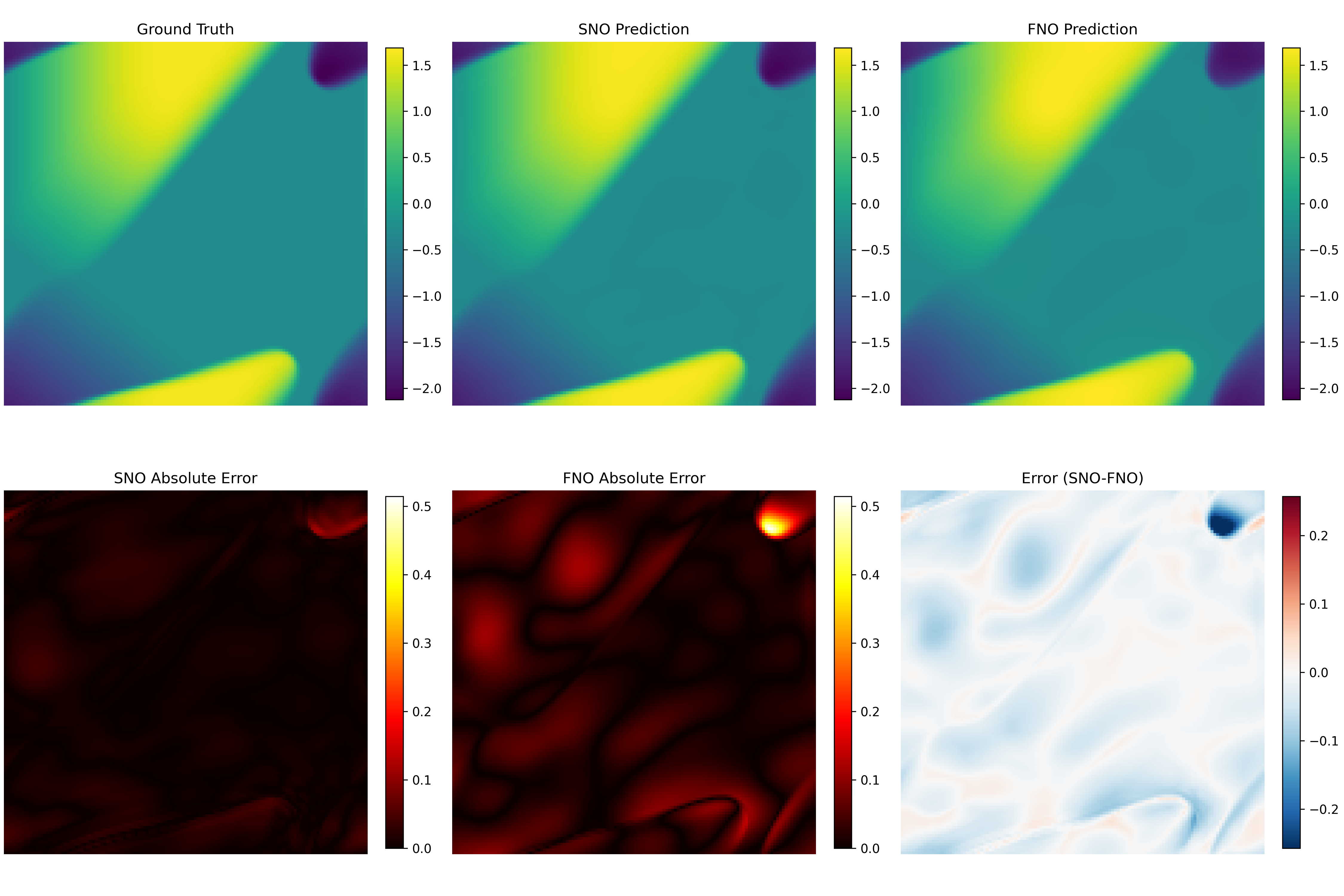}
        \caption{Solution with 128$\times$128 mesh size for multi angle shocks case in T = 220.}
        \label{fig:angle}
    \end{subfigure}
    \hfill 
    \begin{subfigure}[b]{0.49\linewidth}
        \centering
    \includegraphics[width=\textwidth]{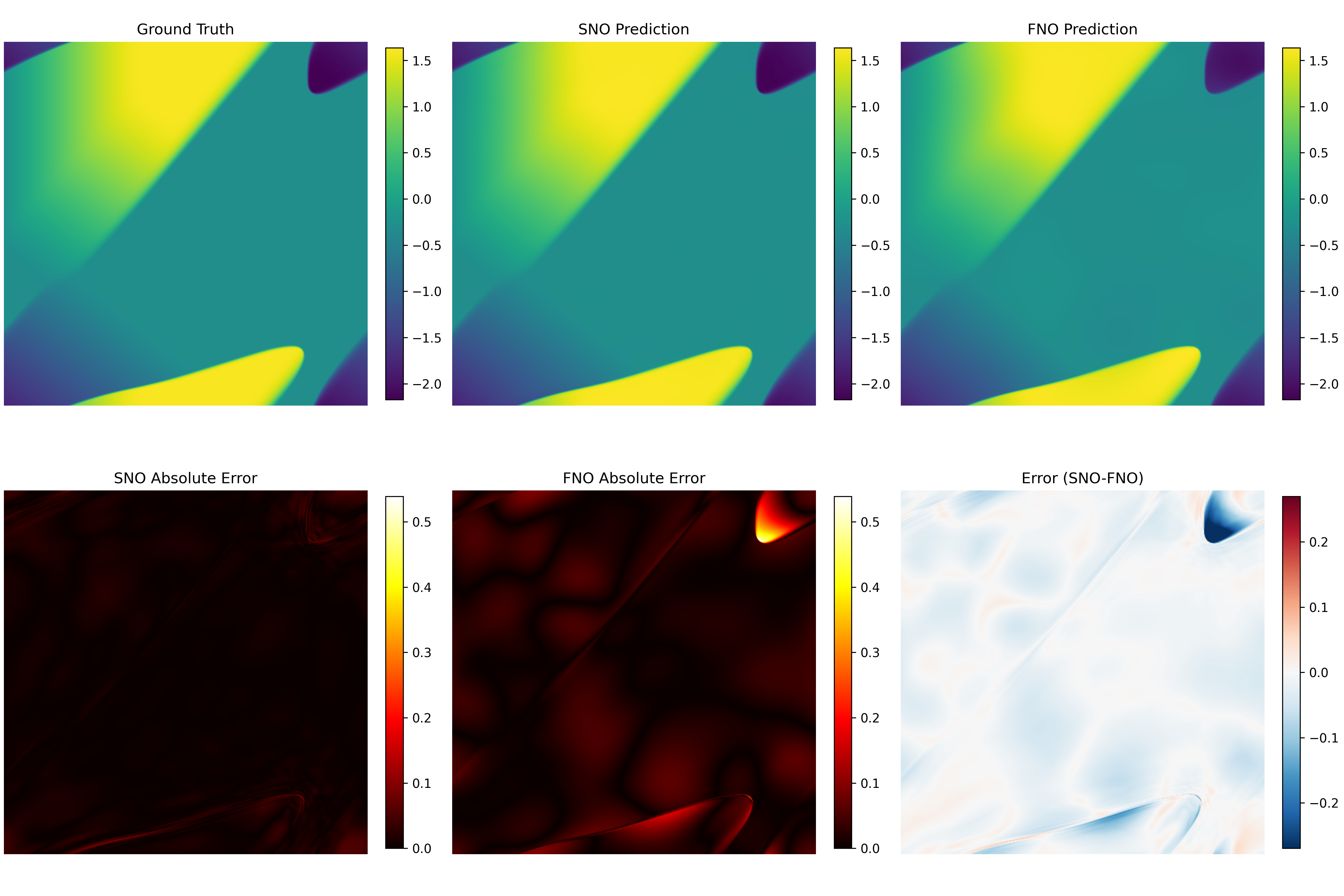}
        \caption{Solution with 512$\times$512 mesh size for multi angle shocks case in T = 220.}
        \label{fig:angle512}
    \end{subfigure}
    \caption{Comparison of the Ground Truth, SNO Prediction, FNO Prediction, their respective Absolute Errors, and the Error Difference between SNO and FNO for a multi angle shocks case  partial differential equation (PDE) solution. }
    \label{fig:mainAngle}
\end{figure}

Figure \ref{fig:mainAngle} presents the true solution of Burgers' equation featuring multiple shock angles. The SNO prediction is visually almost indistinguishable from the ground truth, whereas the FNO prediction exhibits noticeable smoothing and blurring, particularly near shock fronts. This qualitative observation is supported by the absolute error plots: SNO maintains consistently lower error throughout the domain, while FNO shows higher errors, especially in regions with sharp gradients. The error difference colormap further highlights where each model performs better—red regions indicate lower error for SNO, and blue regions indicate lower error for FNO. These results demonstrate that SNO outperforms FNO across nearly the entire domain for both mesh resolution levels (dataset sizes).

\begin{figure}[htbp]
    \centering
    \begin{subfigure}[b]{0.49\linewidth}
        \centering
    \includegraphics[width=\textwidth]{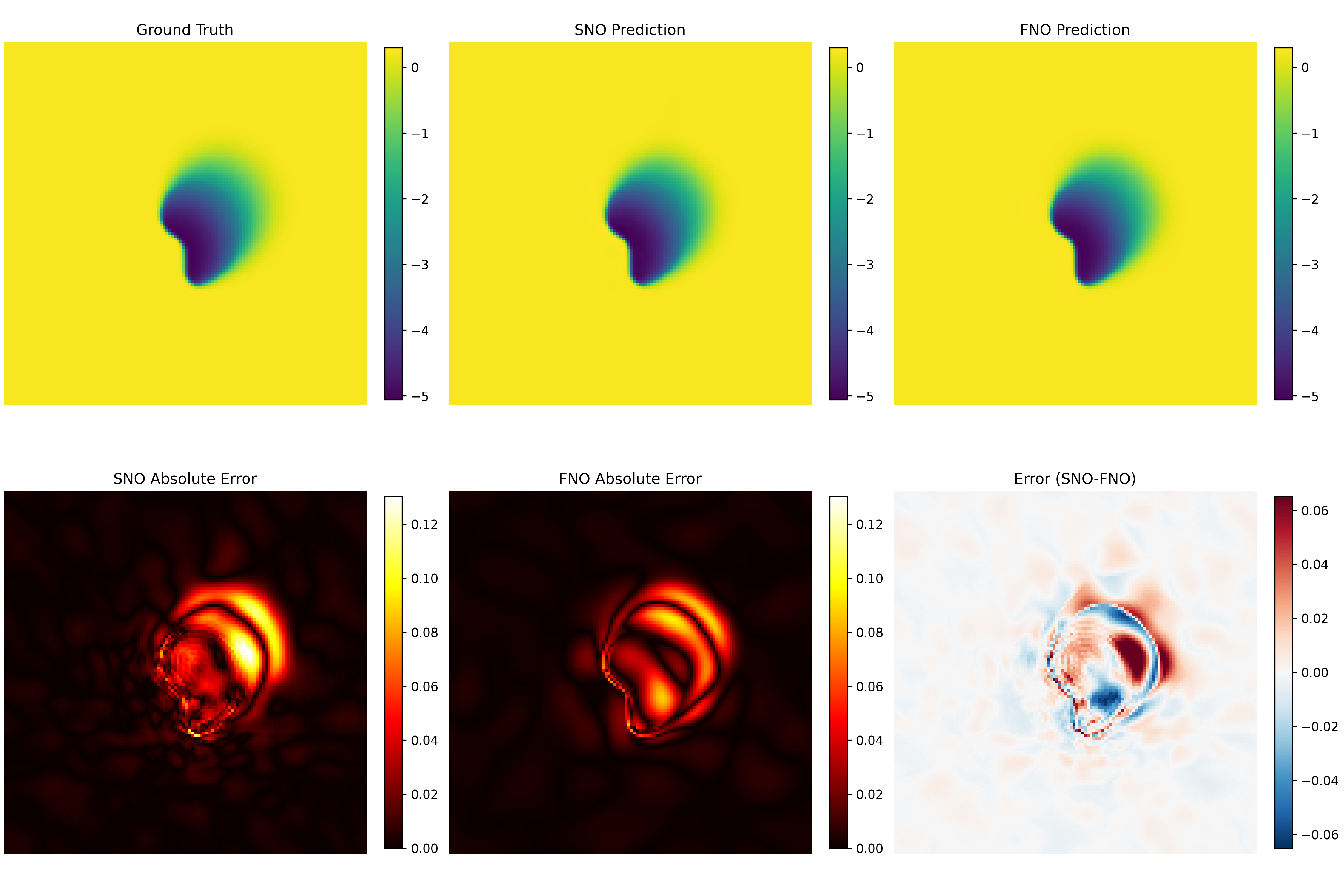}
        \caption{Solution with 128$\times$128 mesh size for spiral shock case in T = 100.}
        \label{fig:spiral}
    \end{subfigure}
    \hfill 
    \begin{subfigure}[b]{0.49\linewidth}
        \centering
    \includegraphics[width=\textwidth]{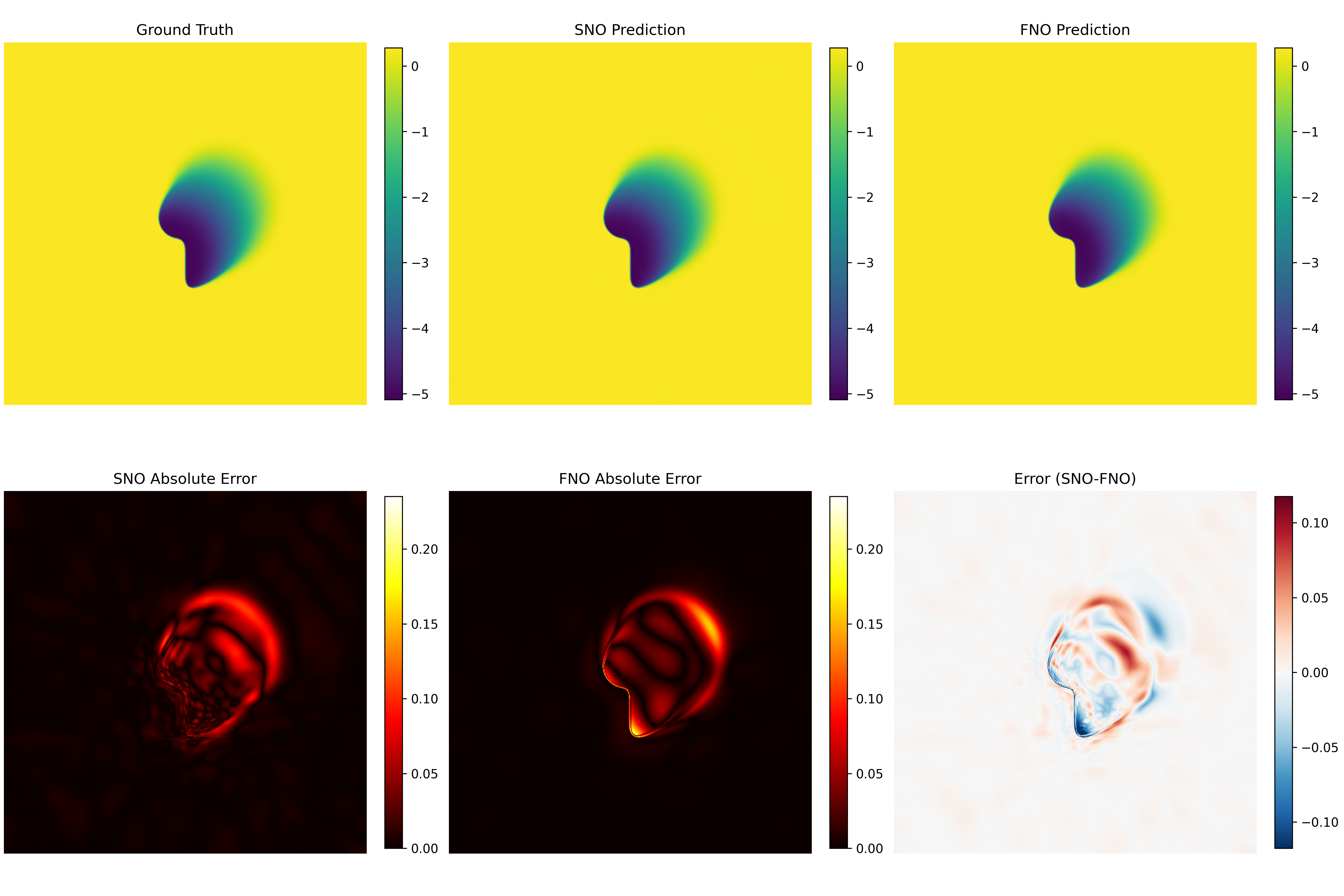}
        \caption{Solution with 512$\times$512 mesh size for spiral shock case in T = 100.}
        \label{fig:spiral512}
    \end{subfigure}
    \caption{Comparison of the Ground Truth, SNO Prediction, FNO Prediction, their respective Absolute Errors, and the Error Difference between SNO and FNO for a spiral shock case partial differential equation (PDE) solution.}
    \label{fig:mainSpiral}
\end{figure}

Figure \ref{fig:mainSpiral} shows the true solution of Burgers' equation for a spiral shock configuration. In this case, while SNO remains superior overall, the performance gap is narrower. The error difference map reveals that SNO excels primarily in regions with strong discontinuities, whereas FNO performs slightly better in smoother areas. This behavior aligns with expectations, given the nature of the two architectures and their respective strengths in handling sharp versus smooth features. Again, FNO is excellent for smooth, periodic, or globally structured problems, while SNO is superior for problems with discontinuities, shocks, or anisotropic features (e.g., Burgers, transport-dominated phenomena, seismic imaging).

\begin{figure}[htbp]
    \centering
    \begin{subfigure}[b]{0.49\linewidth}
        \centering
    \includegraphics[width=\textwidth]{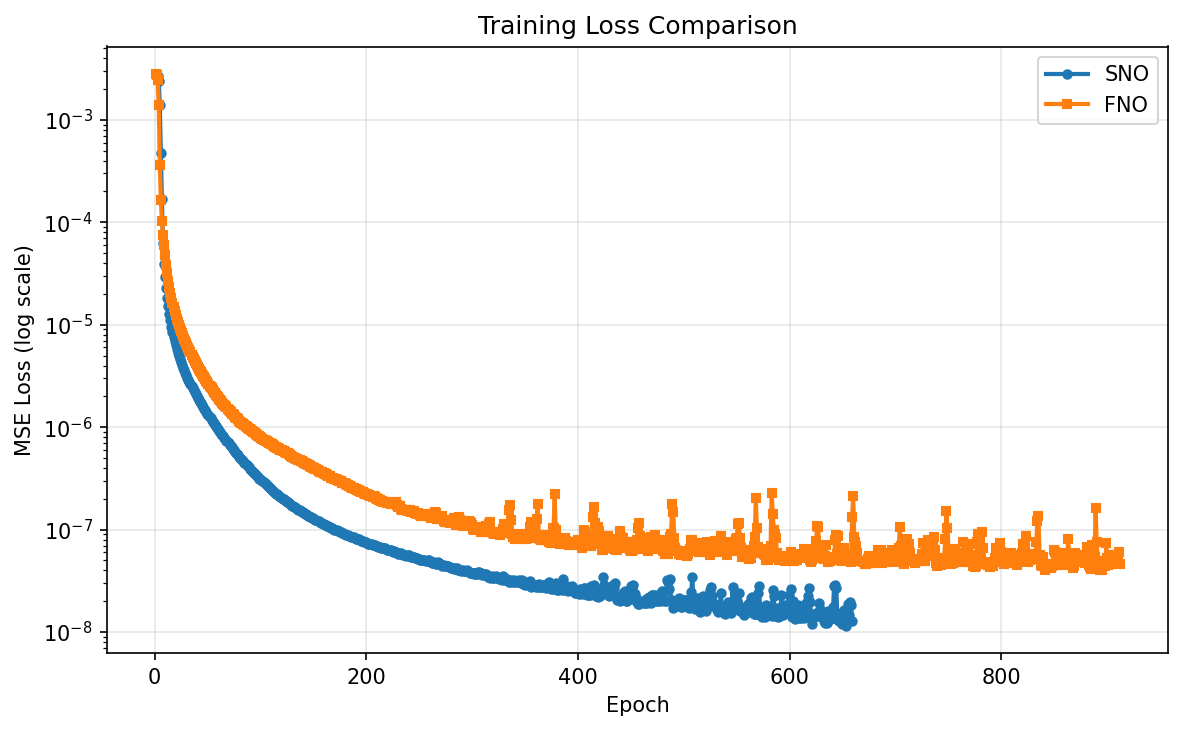}
        \caption{Training loss for mesh 128$\times$128 mesh size for multi angle shocks case.}
        \label{fig:sub1angle}
    \end{subfigure}
    \hfill 
    \begin{subfigure}[b]{0.49\linewidth}
        \centering
    \includegraphics[width=\textwidth]{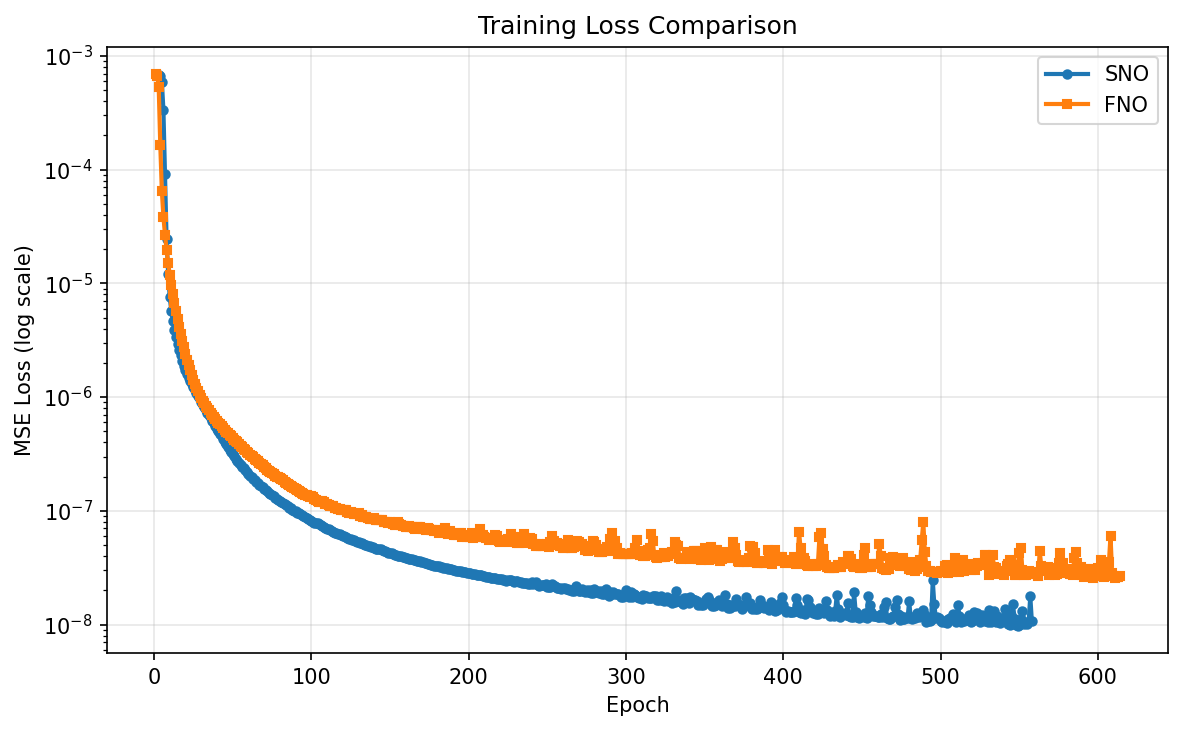}
        \caption{Training loss for mesh 512$\times$512 mesh size for multi angle shocks case.}
        \label{fig:sub2angle}
    \end{subfigure}
    \caption{Training loss comparison for multi angle shocks case between SNO and FNO. The loss (MSE) is plotted on a logarithmic scale against the number of epochs for multi angle shocks case.}
    \label{fig:mainLossAngle}
\end{figure}

According to Figure \ref{fig:mainLossAngle}, SNO achieves a lower MSE than FNO during training, and for large datasets, there is an improvement in the MSE error. Specifically, SNO reaches below $10^{-7}$, while FNO stagnates around $10^{-4}$ to $10^{-5}$. For Figure \ref{fig:mainSpiral}, the result differs. Although SNO remains superior, both are within the same order of magnitude in terms of MSE. This is likely due to the presence of regions—smoother areas—where FNO outperforms SNO.

\begin{figure}[htbp]
    \centering
    \begin{subfigure}[b]{0.49\linewidth}
        \centering
    \includegraphics[width=\textwidth]{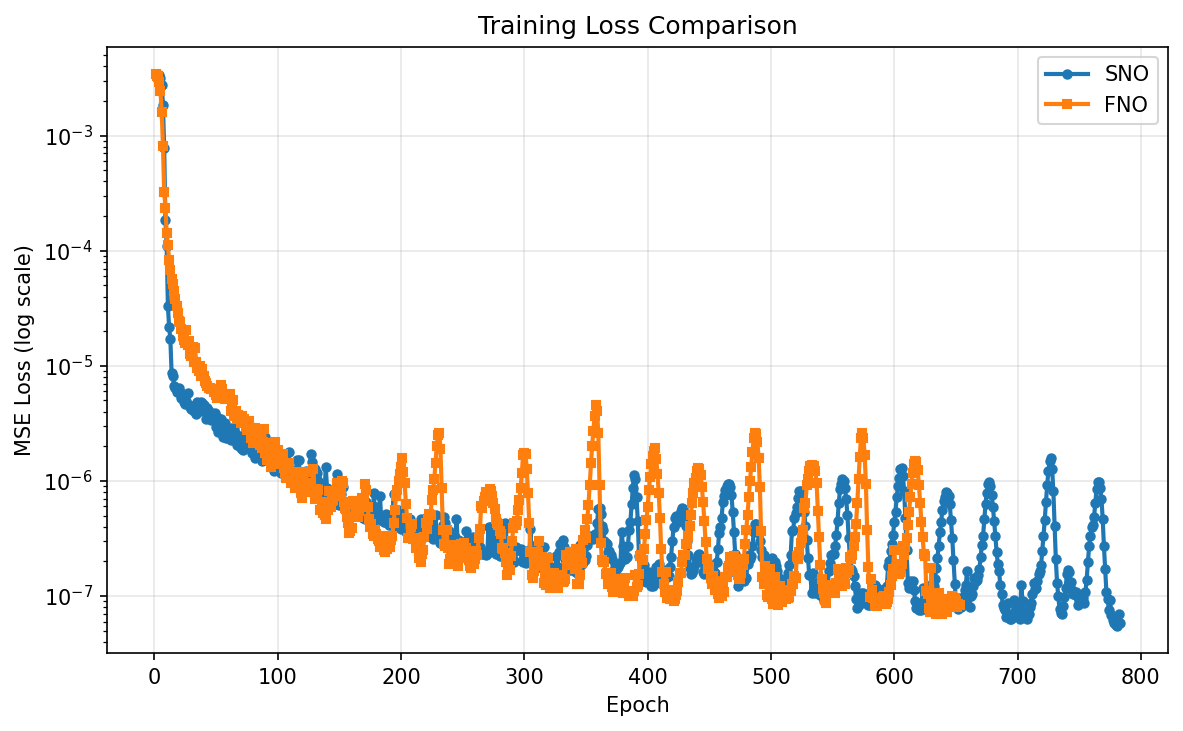}
        \caption{Training loss for mesh 512$\times$512 mesh size for spiral shock case.}
        \label{fig:sub1spiral}
    \end{subfigure}
    \hfill 
    \begin{subfigure}[b]{0.49\linewidth}
        \centering
    \includegraphics[width=\textwidth]{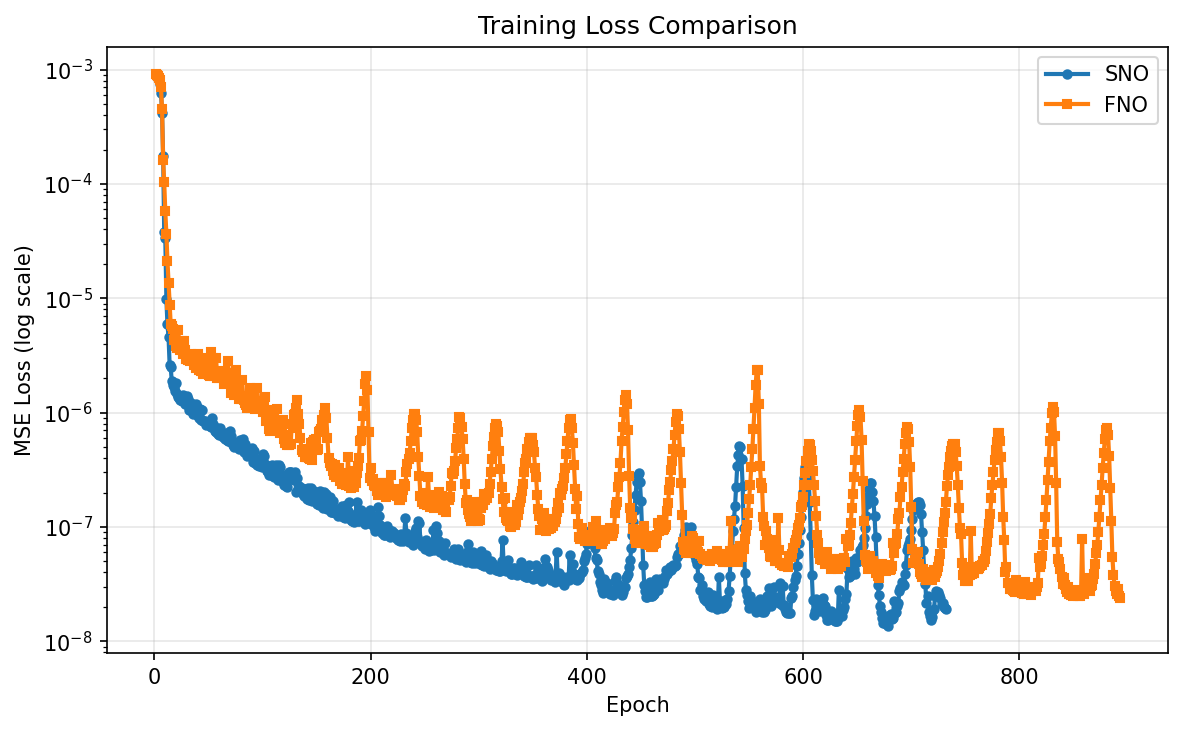}
        \caption{Training loss for mesh 512$\times$512 mesh size for spiral shock case .}
        \label{fig:sub2spiral}
    \end{subfigure}
    \caption{Training loss comparison for spiral shock case between SNO and FNO. The loss (MSE) is plotted on a logarithmic scale against the number of epochs for spiral shock case.}
    \label{fig:mainLossSpiral}
\end{figure}

\subsection{Overall models comparation}

Tables \ref{table128} and \ref{table512} compare the performance of two neural operator models—SNO and FNO—across seven different physical simulation datasets. From these tables, it can be observed that SNO outperforms FNO on six out of the seven datasets across most evaluation metrics. The ratio of the $L_2$ norm is less than one in all but one case, indicating that the errors produced by SNO are typically between 10\% and 60\% of those produced by FNO.

From these results, the advection cases highlight a particularly significant advantage for SNO. Specifically, SNO demonstrates strong improvements in the Structural Similarity Index (SSIM) for the Anisotropic Ridge Advect, Bent Ridge Advect, and Sheared Kelvin–Helmholtz Stripes datasets. This outcome is expected, as FNO is known to struggle with strongly advective problems. In contrast, for the Spiral Shock dataset, FNO exhibits superior performance, possibly because this case contains different spectral characteristics, potentially including higher-frequency features that FNO can better capture.

Furthermore, SNO achieves one to two orders of magnitude lower errors in several cases, with particularly strong performance on the Anisotropic Ridge Advect and Bent Ridge Advect datasets. In terms of Mean Absolute Error (MAE), SNO consistently produces errors approximately 3–10 times smaller than those of FNO. While both models achieve high SSIM values (all above 0.97), SNO reaches near-perfect scores (up to 0.999) on multiple datasets.

This behavior is consistent with the design of the shearlet representation, which is well suited for capturing anisotropic structures and discontinuities. Consequently, the SSIM scores indicate that SNO better preserves important physical features such as shocks, ridges, and interfaces.

Comparing the results presented in Tables \ref{table128} and \ref{table512}, it can be observed that the model performance varies with the dataset size. In particular, both the FNO and SNO models show improved accuracy as the dataset size increases. However, SNO consistently outperforms FNO across the evaluated cases.

\begin{table}[ht]
\centering
\caption{Performance Comparison between SNO and FNO Models for N = 128}
\label{tab:model_comparison1}
\resizebox{\textwidth}{!}{
\begin{tabular}{lccccccccc}
\toprule
\textbf{Dataset} & \textbf{SNO L2} & \textbf{FNO L2} & \textbf{Ratio} & \textbf{SNO MSE} & \textbf{FNO MSE} & \textbf{SNO MAE} & \textbf{FNO MAE} & \textbf{SNO SSIM} & \textbf{FNO SSIM} \\
\midrule
anisotropic ridge advect & $1.58\times10^{-5}$ & $2.01\times10^{-4}$ & 0.0786 & $9.04\times10^{-5}$ & $1.46\times10^{-2}$ & $3.87\times10^{-3}$ & $4.17\times10^{-2}$ & 0.9998 & 0.9862 \\
multi angle shocks & $2.30\times10^{-5}$ & $7.30\times10^{-5}$ & 0.3158 & $1.85\times10^{-4}$ & $1.86\times10^{-3}$ & $9.24\times10^{-3}$ & $2.82\times10^{-2}$ & 0.9973 & 0.9859 \\
bent ridge advect & $1.43\times10^{-5}$ & $1.37\times10^{-4}$ & 0.1040 & $7.35\times10^{-5}$ & $6.79\times10^{-3}$ & $2.94\times10^{-3}$ & $2.67\times10^{-2}$ & 0.9999 & 0.9909 \\
multi orientation texture & $1.27\times10^{-5}$ & $2.15\times10^{-5}$ & 0.5904 & $4.98\times10^{-6}$ & $1.43\times10^{-5}$ & $1.71\times10^{-3}$ & $3.66\times10^{-3}$ & 0.9994 & 0.9960 \\
spiral shock & $4.18\times10^{-5}$ & $3.40\times10^{-5}$ & 1.2278 & $3.27\times10^{-4}$ & $2.17\times10^{-4}$ & $7.22\times10^{-3}$ & $6.09\times10^{-3}$ & 0.9921 & 0.9950 \\
polygonal shock & $1.29\times10^{-5}$ & $3.52\times10^{-5}$ & 0.3660 & $3.13\times10^{-5}$ & $2.34\times10^{-4}$ & $4.25\times10^{-3}$ & $1.21\times10^{-2}$ & 0.9987 & 0.9900 \\
sheared kelvin helmotz & $2.74\times10^{-5}$ & $1.07\times10^{-4}$ & 0.2551 & $1.04\times10^{-5}$ & $1.60\times10^{-4}$ & $2.49\times10^{-3}$ & $1.04\times10^{-2}$ & 0.9986 & 0.9729 \\
\bottomrule
\label{table128}
\end{tabular}
}
\end{table}

\begin{table}[ht]
\centering
\caption{Performance Comparison between SNO and FNO Models for N = 512}
\label{tab:model_comparison2}
\resizebox{\textwidth}{!}{
\begin{tabular}{lccccccccc}
\toprule
\textbf{Dataset} & \textbf{SNO L2} & \textbf{FNO L2} & \textbf{Ratio} & \textbf{SNO MSE} & \textbf{FNO MSE} & \textbf{SNO MAE} & \textbf{FNO MAE} & \textbf{SNO SSIM} & \textbf{FNO SSIM} \\
\midrule
anisotropic ridge advect & $2.41\times10^{-6}$ & $3.97\times10^{-5}$ & 0.0608 & $3.36\times10^{-5}$ & $9.08\times10^{-3}$ & $2.11\times10^{-3}$ & $4.60\times10^{-2}$ & 0.9999 & 0.9583 \\
multi angle shocks & $3.93\times10^{-6}$ & $1.48\times10^{-5}$ & 0.2650 & $9.08\times10^{-5}$ & $1.29\times10^{-3}$ & $6.25\times10^{-3}$ & $2.17\times10^{-2}$ & 0.9990 & 0.9942 \\
bent ridge advect & $2.53\times10^{-6}$ & $2.52\times10^{-5}$ & 0.1004 & $3.69\times10^{-5}$ & $3.67\times10^{-3}$ & $2.43\times10^{-3}$ & $2.24\times10^{-2}$ & 0.9999 & 0.9780 \\
multi orientation texture & $4.09\times10^{-6}$ & $4.21\times10^{-6}$ & 0.9699 & $8.38\times10^{-6}$ & $8.91\times10^{-6}$ & $2.25\times10^{-3}$ & $2.46\times10^{-3}$ & 0.9986 & 0.9988 \\
spiral shock & $8.36\times10^{-6}$ & $8.49\times10^{-6}$ & 0.9852 & $2.16\times10^{-4}$ & $2.22\times10^{-4}$ & $5.84\times10^{-3}$ & $4.42\times10^{-3}$ & 0.9939 & 0.9942 \\
polygonal shock & $3.52\times10^{-6}$ & $8.18\times10^{-6}$ & 0.4308 & $4.10\times10^{-5}$ & $2.21\times10^{-4}$ & $4.19\times10^{-3}$ & $1.04\times10^{-2}$ & 0.9987 & 0.9934 \\
sheared kelvin helmotz & $4.18\times10^{-6}$ & $2.31\times10^{-5}$ & 0.1812 & $3.21\times10^{-5}$ & $9.77\times10^{-4}$ & $4.44\times10^{-3}$ & $2.65\times10^{-2}$ & 0.9989 & 0.9677 \\
\bottomrule
\label{table512}
\end{tabular}
}
\end{table}

This table convincingly demonstrates SNO's superiority for most tested physical systems, though the spiral shock exception warrants further investigation into the architectural differences between these neural operators.

\section{Conclusions}\label{sec:conclusion}

Neural operators have emerged as a powerful data-driven framework for learning solution operators of parametric partial differential equations (PDEs). Several neural operator architectures have been proposed, each tailored to specific classes of problems and data structures. Among the most successful approaches is the Fourier Neural Operator (FNO), which relies on a global Fourier representation to approximate solution operators efficiently.

FNO encounters difficulties when applied to problems characterized by sharp discontinuities or strongly advective dynamics. In particular, advection-dominated systems and conservation laws often exhibit shocks and localized structures that are not well captured by global Fourier bases. Similar limitations are well known in classical spectral methods applied to nonlinear conservation laws.

To address these challenges, we introduce the Shearlet Neural Operator (SNO). The proposed approach replaces the Fourier transform with a shearlet transform within the neural operator framework. The Shearlet Transform was specifically developed to represent anisotropic features, edges, and sharp interfaces in multiscale data. By incorporating this representation, the resulting neural operator architecture becomes better suited for physical systems characterized by directional structures, discontinuities, and multiscale anisotropy.

Overall, the SNO advances scientific machine learning by combining the operator-learning paradigm with the multiscale and directional properties of shearlet representations. In our benchmark experiments, this integration leads to improved generalization and predictive accuracy compared with FNO. Specifically, SNO outperforms FNO on six out of seven datasets across most evaluation metrics. In the remaining case, the performance of both models is comparable due to the smoother behavior of the spiral structure in a diffusion-dominated region.

It is also important to note that increasing the size of the training dataset generally improves the performance of SNO across most cases, although the achievable accuracy remains bounded by the representational capacity of the model parameters. These results suggest that SNO represents a promising direction for learning solution operators of parametric PDEs, particularly in regimes characterized by anisotropy, shocks, or other localized features.

\section*{Acknowledgments}
This work was supported by the Serrapilheira Institute (grant number Serra – R-2111-39718)

\bibliographystyle{plain}

\begin{thebibliography}{10}

\bibitem{BEAULIEU2009105}
Christian Beaulieu.
\newblock Chapter 6 - the biological basis of diffusion anisotropy.
\newblock In Heidi Johansen-Berg and Timothy~E.J. Behrens, editors, {\em Diffusion MRI}, pages 105--126. Academic Press, San Diego, 2009.

\bibitem{cao2020understandingspectralbiasdeep}
Yuan Cao, Zhiying Fang, Yue Wu, Ding-Xuan Zhou, and Quanquan Gu.
\newblock Towards understanding the spectral bias of deep learning, 2020.

\bibitem{crank1975mathematics}
John Crank.
\newblock {\em The Mathematics of Diffusion}.
\newblock Clarendon Press, Oxford, UK, 2nd edition, 1975.

\bibitem{GottliebShu1998}
Sigal Gottlieb and Chi-Wang Shu.
\newblock Total variation diminishing runge--kutta schemes.
\newblock {\em Mathematics of Computation}, 67(221):73--85, 1998.

\bibitem{kutyniok2006shearlets}
Kanghui Guo, Gitta Kutyniok, and Demetrio Labate.
\newblock Shearlets: Sparse multidimensional representations using anisotropic dilation and shear operators.
\newblock In {\em Wavelets and Splines}, 05 2005.

\bibitem{HU2025107649}
Yifan Hu, Weimin Zhang, Fukang Yin, and Jianping Wu.
\newblock Hmgno: Hybrid multigrid neural operator with low-order numerical solver for partial differential equations.
\newblock {\em Neural Networks}, 190:107649, 2025.

\bibitem{isakov2013inverse}
V.~Isakov.
\newblock {\em Inverse Problems for Partial Differential Equations}.
\newblock Applied Mathematical Sciences. Springer New York, 2013.

\bibitem{John_Xu_2020}
Zhi-Qin John~Xu, Yaoyu Zhang, Tao Luo, Yanyang Xiao, and Zheng Ma.
\newblock Frequency principle: Fourier analysis sheds light on deep neural networks.
\newblock {\em Communications in Computational Physics}, 28(5):1746–1767, June 2020.

\bibitem{10.5555/3648699.3648788}
Nikola Kovachki, Zongyi Li, Burigede Liu, Kamyar Azizzadenesheli, Kaushik Bhattacharya, Andrew Stuart, and Anima Anandkumar.
\newblock Neural operator: learning maps between function spaces with applications to pdes.
\newblock {\em J. Mach. Learn. Res.}, 24(1), January 2023.

\bibitem{kutyniok2012shearlets}
Gitta Kutyniok and Demetrio Labate, editors.
\newblock {\em Shearlets}.
\newblock Applied and Numerical Harmonic Analysis. Birkh{\"a}user, Boston, MA, 1 edition, 2012.

\bibitem{Kutyniok_2016}
Gitta Kutyniok, Wang-Q Lim, and Rafael Reisenhofer.
\newblock Shearlab 3d: Faithful digital shearlet transforms based on compactly supported shearlets.
\newblock {\em ACM Transactions on Mathematical Software}, 42(1):1–42, January 2016.

\bibitem{LeVeque_2002}
Randall~J. LeVeque.
\newblock {\em Finite Volume Methods for Hyperbolic Problems}.
\newblock Cambridge Texts in Applied Mathematics. Cambridge University Press, 2002.

\bibitem{li2021fourierneuraloperatorparametric}
Zongyi Li, Nikola Kovachki, Kamyar Azizzadenesheli, Burigede Liu, Kaushik Bhattacharya, Andrew Stuart, and Anima Anandkumar.
\newblock Fourier neural operator for parametric partial differential equations, 2021.

\bibitem{li2023physicsinformedneuraloperatorlearning}
Zongyi Li, Hongkai Zheng, Nikola Kovachki, David Jin, Haoxuan Chen, Burigede Liu, Kamyar Azizzadenesheli, and Anima Anandkumar.
\newblock Physics-informed neural operator for learning partial differential equations, 2023.

\bibitem{liu2021multiscaledeeponetnonlinearoperators}
Lizuo Liu and Wei Cai.
\newblock Multiscale deeponet for nonlinear operators in oscillatory function spaces for building seismic wave responses, 2021.

\bibitem{Lu_2021}
Lu~Lu, Pengzhan Jin, Guofei Pang, Zhongqiang Zhang, and George~Em Karniadakis.
\newblock Learning nonlinear operators via deeponet based on the universal approximation theorem of operators.
\newblock {\em Nature Machine Intelligence}, 3(3):218–229, March 2021.

\bibitem{pmlr-v190-luo22a}
Tao Luo, Zheng Ma, Zhiwei Wang, Zhiqin John~Xu, and Yaoyu Zhang.
\newblock An upper limit of decaying rate with respect to frequency in linear frequency principle model.
\newblock In Bin Dong, Qianxiao Li, Lei Wang, and Zhi-Qin~John Xu, editors, {\em Proceedings of Mathematical and Scientific Machine Learning}, volume 190 of {\em Proceedings of Machine Learning Research}, pages 205--214. PMLR, 15--17 Aug 2022.

\bibitem{luo2019theoryfrequencyprinciplegeneral}
Tao Luo, Zheng Ma, Zhi-Qin~John Xu, and Yaoyu Zhang.
\newblock Theory of the frequency principle for general deep neural networks, 2019.

\bibitem{doi:10.1137/21M1444400}
Tao Luo, Zheng Ma, Zhi-Qin~John Xu, and Yaoyu Zhang.
\newblock On the exact computation of linear frequency principle dynamics and its generalization.
\newblock {\em SIAM Journal on Mathematics of Data Science}, 4(4):1272--1292, 2022.

\bibitem{10.1098/rspa.2016.0117}
James~C. McWilliams.
\newblock Submesoscale currents in the ocean.
\newblock {\em Proceedings of the Royal Society A: Mathematical, Physical and Engineering Sciences}, 472(2189):20160117, 05 2016.

\bibitem{doi:10.1137/1018130}
Robert~M. Miura.
\newblock Linear and nonlinear waves (g. b. whitham).
\newblock {\em SIAM Review}, 18(4):776--781, 1976.

\bibitem{rahaman2019spectralbiasneuralnetworks}
Nasim Rahaman, Aristide Baratin, Devansh Arpit, Felix Draxler, Min Lin, Fred~A. Hamprecht, Yoshua Bengio, and Aaron Courville.
\newblock On the spectral bias of neural networks, 2019.

\bibitem{RAISSI2019686}
M.~Raissi, P.~Perdikaris, and G.E. Karniadakis.
\newblock Physics-informed neural networks: A deep learning framework for solving forward and inverse problems involving nonlinear partial differential equations.
\newblock {\em Journal of Computational Physics}, 378:686--707, 2019.

\bibitem{fluids9030052}
Mae Sementilli, Rozie Zangeneh, and James Chen.
\newblock Influence of cross perturbations on turbulent kelvin–helmholtz instability.
\newblock {\em Fluids}, 9(3), 2024.

\bibitem{smith2024uncertainty}
R.C. Smith.
\newblock {\em Uncertainty Quantification: Theory, Implementation, and Applications, Second Edition}.
\newblock Computational science and engineering. Society for Industrial and Applied Mathematics, 2024.

\bibitem{81d297}
T.~J. Sullivan.
\newblock {\em {Introduction to Uncertainty Quantification}}, volume~63 of {\em Texts in Applied Mathematics}.
\newblock Springer, 2015.

\bibitem{wiatowski2018energypropagationdeepconvolutional}
Thomas Wiatowski, Philipp Grohs, and Helmut Bölcskei.
\newblock Energy propagation in deep convolutional neural networks, 2018.

\bibitem{xu2024overviewfrequencyprinciplespectralbias}
Zhi-Qin~John Xu, Yaoyu Zhang, and Tao Luo.
\newblock Overview frequency principle/spectral bias in deep learning, 2024.

\bibitem{xu2019trainingbehaviordeepneural}
Zhi-Qin~John Xu, Yaoyu Zhang, and Yanyang Xiao.
\newblock Training behavior of deep neural network in frequency domain, 2019.

\bibitem{you2024mscalefnomultiscalefourierneural}
Zhilin You, Zhenli Xu, and Wei Cai.
\newblock Mscalefno: Multi-scale fourier neural operator learning for oscillatory function spaces, 2024.

\bibitem{sadiq2020shearletcnn}
Chaymae Ziani and Abdelalim Sadiq.
\newblock Sh-cnn: Shearlet convolutional neural network for gender classification.
\newblock {\em Advances in Science, Technology and Engineering Systems Journal}, 5:1328--1334, 12 2020.

\end{thebibliography}

\end{document}